\documentclass[preprint,12pt,authoryear]{elsarticle}

\usepackage{amsmath,amssymb}
\usepackage{graphicx}
\usepackage{tabularx}
\usepackage{booktabs}
\usepackage{multirow}

\usepackage{siunitx}
\usepackage{subcaption}
\usepackage{rotating}
\usepackage{array}
\usepackage{fontawesome5}
\usepackage{tikz}
\usepackage{float}
\usepackage[table]{xcolor}

\usepackage{hyperref}
\usepackage{adjustbox}
\usepackage{xspace}

\usetikzlibrary{arrows.meta, positioning, calc}

\usepackage{pgfplots}
\pgfplotsset{compat=1.18}

\definecolor{axP}{HTML}{7F77DD}\definecolor{axPbg}{HTML}{EEEDFE}\definecolor{axPdk}{HTML}{26215C}
\definecolor{axT}{HTML}{1D9E75}\definecolor{axTbg}{HTML}{E1F5EE}\definecolor{axTdk}{HTML}{04342C}
\definecolor{axC}{HTML}{D85A30}\definecolor{axCbg}{HTML}{FAECE7}\definecolor{axCdk}{HTML}{4A1B0C}

\newcommand{\mgha}{\,Mg/ha\xspace}

\journal{Remote Sensing of Environment}

\begin{document}

\begin{frontmatter}

\title{Multi-Modal, Multi-Temporal Geospatial Foundation Models Improve Global Above-Ground Biomass Estimation}

\author[label1,label2]{Ghjulia Sialelli\corref{cor1}}
\ead{gsialelli@ethz.ch}
\cortext[cor1]{Corresponding author.}
\affiliation[label1]{organization={Photogrammetry and Remote Sensing},
             addressline={ETH Zurich},
             city={Zurich},
             postcode={8049},
             country={Switzerland}}
\affiliation[label2]{organization={ETH AI Center},
             city={Zurich},
             postcode={8092},
             country={Switzerland}}
\affiliation[label3]{organization={EcoVision Lab, Department of Mathematical Modeling and Machine Learning},
            addressline={University of Zurich},
           city={Zurich},
             postcode={8057},
          country={Switzerland}}
\author[label1]{Linus Scheibenreif}
\author[label3]{Jan Dirk Wegner}
\author[label1,label2]{Konrad Schindler}

\begin{abstract}
Accurate estimation of above-ground biomass (AGB) from satellite imagery is essential for the large-scale monitoring of carbon stocks, yet it remains a challenging regression task at global scale. Geospatial Foundation Models (GFMs) have recently emerged as a promising machine learning paradigm to derive general-purpose representations from Earth observation (EO) data, but their utility for quantitative regression tasks like biomass estimation remains largely unexplored, as most benchmarks focus on classification and segmentation. Here, we benchmark 13 GFMs for above-ground biomass estimation, specifically GEDI L4A biomass values. None of them was designed for this task, and they differ widely in input modalities, use of time series and pre-training objectives. We compare them against a fully supervised state-of-the-art (SOTA) baseline trained on A Global-scale Biomass Dataset (AGBD), a machine learning-ready benchmark spanning diverse biomes and geographies. Most of the GFMs evaluated substantially underperform with respect to the supervised baseline, whether frozen or fine-tuned; only TESSERA and AlphaEarth Foundations (AEF) features with certain prediction heads yield a better performance. Our SOTA model trained on AEF embeddings and selected raw EO features achieves the best overall result (a root mean squared error (RMSE) of $50.79$\mgha vs. $53.73$\mgha on AGBD features). This feature combination also generalizes better than AGBD features: it reduces RMSE by up to $25\%$ in selected regions excluded from training, and, trained on single-year data alone, it outperforms the model trained on multi-year AGBD features. With labels from only 11 regions around the world, the AEF-based model also performs on par with the globally calibrated ESA CCI biomass product on independent reference plots (RMSE $44.79$\mgha vs. $45.77$\mgha). These findings indicate that GFM features can be highly informative for biomass estimation, when the underlying model is trained on rich multi-modal data. Distributed as accessible embedding layers, they lower the computational barrier to state-of-the-art biomass estimation for practitioners.
\end{abstract}


\end{frontmatter}

\section{Introduction}
\label{sec:introduction}

Above-ground biomass (AGB) quantifies the carbon stored in vegetation, making it a key variable for both climate science and ecology, as forests are among the largest terrestrial carbon sinks \citep{pan2024} and shelter much of the world's biodiversity \citep{FAOUNEP2020}. Accurate AGB maps are therefore essential for carbon accounting, climate policy, and conservation, yet maps that are simultaneously high-resolution, global, and regularly updated remain scarce. One reason is that biomass is exceptionally expensive to observe. At the level of an individual tree, biomass can only be measured directly by felling and weighing it. Field surveys avoid this through \emph{allometric equations}, which infer biomass from simple measurements of a standing tree \citep{picard2012}. Following protocols such as UN-REDD \citep{moges2010}, inventories record a tree's diameter, height, and species, then apply a suitable allometric equation. The choice of equation is consequential, as databases like GlobAllomeTree list thousands of equations \citep{henry2013} that are continually refined \citep{chave2014,jucker2017}. However, field data is slow and costly to collect and covers only a small, uneven share of the world's forests. LiDAR offers a way past sparse field plots, recording forest structure as dense 3D point clouds \citep{disney2018}, and can be acquired from tripods, vehicles, or aircraft \citep{kangas2018}. Terrestrial and mobile scanning capture fine details under the canopy, yet still depend on fieldwork \citep{holvoet2025}. Airborne acquisition maps wide areas and now feeds national inventories \citep{nilsson2017,monnet2016}, but is expensive and slow to scale. These limits push the field toward the use of satellite remote sensing. 

Spaceborne missions provide free, global, and frequently repeated imagery, from passive optical sensors such as Sentinel-2 \citep{drusch2012} to active SAR sensors \citep{moreira2013}. The two are complementary: optical sensors capture fine-grained spectral detail of the canopy but require clear skies and saturate over dense forest, whereas SAR penetrates clouds and is less prone to saturation, especially at longer L- and P-band wavelengths \citep{Imhoff1995,Naidoo2015}. These missions are general-purpose EO platforms, providing continuous signals that correlate with biomass without being tailored to it. A complementary class of missions has instead been designed with vegetation structure in mind: GEDI's spaceborne LiDAR \citep{dubayah2020} yields sparse AGB estimates derived from observed canopy structure \citep{GEDIL4A}, and missions such as ESA's Biomass\footnote{\url{https://earth.esa.int/eogateway/catalog/biomass-level-2b}} and NISAR\footnote{\url{https://science.nasa.gov/mission/nisar/}} extend this further. Machine learning (ML) has become the standard tool for turning this ever-growing archive of EO data into actionable information, progressing from hand-crafted features and shallow classifiers to deep neural networks that learn representations directly from imagery~\citep{zhu2017deep}. This has enabled the large-scale estimation of forest properties such as canopy height \citep{lang_canopy,Potapov2021,Pauls2024,TOLAN2024113888} and above-ground biomass \citep{schwartz2023forms,agbd2024}, by training neural networks on satellite imagery calibrated against GEDI or airborne LiDAR reference data. 

Several limitations constrain current approaches. First, satellite-derived AGB estimates exhibit a persistent bias, overestimating at the low end of the biomass range and underestimating at the high end \citep{avitabile2016,RodrguezVeiga2019}. This is attributed to the aforementioned saturation of the satellite signal over closed canopies and to the long-tailed distribution of biomass values. Second, the GEDI labels are unevenly distributed: GEDI's orbit is confined to latitudes below approximately 51.6$^{\circ}$, excluding boreal forests; and its footprint-level estimates are calibrated against field data, inheriting documented geographical biases \citep{Duncanson2022,Pascual2023}. 

Geospatial Foundation Models (GFMs), the latest development in the steadily advancing line of ML methods for remote sensing, have been proposed to address problems of this kind \citep{prithvi,scalemae,spectralgpt,feng2025tesseratemporalembeddingssurface}. These large neural networks are pre-trained on large quantities of unlabeled satellite imagery in a self-supervised manner, i.e., using proxy objectives that can be constructed from the data alone. The hope is that this yields broadly applicable, general representations that can serve as a basis for different downstream tasks, while requiring only a minimal quantity of labels to adapt to the task at hand. This is particularly relevant for biomass estimation: these multi-modal, multi-temporal representations may carry rich information, and the unconstrained geographical distribution of the pre-training data may enable generalization to regions where labels are sparse or unreliable. Once pre-trained, satellite imagery is passed through the GFMs' encoders to produce dense numerical representations, known as embeddings, that condense complex spectral and spatial patterns into a low-dimensional vector space. Task-specific heads, ranging from a simple linear layer to multi-million parameter models, are trained on top. In this setting, GFMs can be adapted to downstream tasks via several strategies. Full-parameter fine-tuning updates the entire encoder, but is often impractical due to the massive parameter count of these models and the scarcity of labeled data. Parameter-efficient methods such as Low-Rank Adaptation (LoRA) \citep{hu2022lowrank} have recently emerged to reduce this cost by updating only a small subset of parameters. More commonly, practitioners treat the encoder as a frozen feature extractor, the encoder parameters remaining unchanged \citep{Bourriz2026,Hossain2026,Kaushik2026}. In any case, generating such embeddings over large areas requires significant computational overhead for the downstream user, including the ingestion of massive EO datasets and access to high-end GPU resources. To bypass this, some GFMs \citep{alphaearth,feng2025tesseratemporalembeddingssurface} distribute pre-computed, per-pixel embeddings as a standalone product rather than, or in addition to, releasing model parameters, effectively turning the model's learned knowledge into a ready-to-use geospatial layer. 

To leverage the strengths of GFMs, biomass practitioners must choose which model to use, whether and how to fine-tune it, or whether to rely on pre-computed embeddings, with little evidence on which approach works best. This choice is not trivial, as GFMs differ widely in input modalities, use of time series and pre-training objectives, and are developed as general-purpose models rather than for AGB estimation. Several GFM evaluation frameworks have emerged in recent years (GEO-Bench \citep{lacoste2023geobench}, PANGAEA \citep{pangaea2024}, PhilEO \citep{fibaek2024PhilEO}, FoMo \citep{bountos2023fomo}, SustainBench \citep{yeh2021sustainbench}, MMEarth-Bench \citep{gordon2026mmearthbench}, Copernicus-Bench \citep{wang2025unifiedcopernicusfoundationmodel}), but most of the benchmark tasks are classification or segmentation problems, and continuous-valued regression tasks such as biomass estimation remain comparatively rare \citep{nascetti2023biomassters,gordon2026mmearthbench}. Beyond benchmarks, several studies have combined GFM representations with AGB reference data, with mixed results. They performed slightly below a supervised baseline trained from scratch in Brazil, though with far fewer trainable parameters \citep{muszynski2024}, and below conventional spectral indices in tropical Andean forests, albeit over a small study area and with low-capacity models \citep{RojasLucero2026}, but reached accuracies comparable to LiDAR-based studies in subtropical forests \citep{Jin2026}. Calibrated with national forest inventory plots, they also outperformed GEDI-based biomass estimates in Spain’s largest forest region \citep{Pascual2026}. Between benchmarks that rarely include biomass regression and studies that are mostly regional and tend to focus on a single model, the field still lacks a comprehensive benchmark of GFMs for biomass estimation, covering both frozen and fine-tuned models and their generalization across space and time.

Our contributions are as follows:
\begin{enumerate}
    \item We extend the PANGAEA evaluation framework with a globally distributed biomass regression task.
    \item We benchmark 13 GFMs on A Global-scale Biomass Dataset (AGBD) \citep{agbd2024}, and compare them against a state-of-the-art fully supervised model.
    \item We design and analyze geographical and temporal generalization experiments.
    \item We further compare our best model's performance with the widely used ESA CCI biomass maps \citep{ESACCI} on an independent reference dataset, AGBref \citep{AGBref}.
\end{enumerate}

Our central finding is that, for biomass regression, most GFMs (whether frozen or LoRA fine-tuned) underperform with respect to a fully supervised baseline. Only AEF and TESSERA embeddings surpassed the SOTA supervised model. Our results suggest that the promise of foundation models for environmental monitoring is best realized when a model trained on rich, multi-temporal, multi-modal data is distributed as an accessible embedding layer.

\section{Data}
\label{sec:data}

AGBD \citep{agbd2024} is a large-scale, ML-ready dataset for AGB estimation from EO data, comprising approximately 16 million patches distributed globally. Each sample pairs AGB values from NASA's GEDI L4A mission \citep{GEDIL4A}, which we use as reference, with co-located $25\times25$-pixel patches, which we use as input, from multiple sources:
\begin{itemize}
    \item Sentinel-2 L2A \citep{drusch2012}: $12$ bands of multispectral optical imagery at 10m, 20m and 60m resolutions.
    \item ALOS-2 PALSAR-2 \citep{rosenqvist2007}: L-band SAR backscatter, at 25m resolution.
    \item Ancillary features: the ALOS Global Digital Surface Model \citep{isprs-annals-II-4-71-2014} at 30m resolution; the Copernicus Global Land Service Dynamic Land Cover map \citep{ESALC} for 2019 at 100m resolution; a Canopy height map \citep{lang_canopy} at 10m resolution; and geographical coordinates. 
\end{itemize}

Each patch is built around a single GEDI footprint of about 25m in diameter. As the laser energy decreases rapidly away from the footprint center, the central pixel of the patch receives the largest share of the signal \citep{pauls2026echosat}, and the footprint’s AGB value is assigned to that pixel \citep{agbd2024}. The dataset spans diverse biomes across all vegetated continents, with GEDI footprints collected between 2019 and 2020. Specifically, the AGBD covers the following regions: California (USA), Cuba, French Guiana, Paraguay, Austria, Greece, Ghana, Tanzania, New Zealand, Nepal, and the Shaanxi province (China). The geographic and temporal breadth of AGBD makes it uniquely suited to evaluate generalization capabilities. We use the original AGBD split, which is defined at the level of 100km by 100km Sentinel-2 tiles: in each region, 65\%, 15\% and 20\% of the tiles are used for training, validation and testing, respectively, so that test samples always come from tiles unseen during training.

The complete AGBD, with its $\approx$16 million samples, poses computational challenges for the evaluation of GFMs within the PANGAEA framework. To enable efficient preliminary benchmarking, we construct \textbf{AGBD Lite}, a representative subset of the full dataset (or \textbf{AGBD Full}) made of $\approx$600,000 samples. We also construct a multi-temporal extension, called \textbf{AGBD-MT Lite}, with 3 Sentinel-2 timesteps per GEDI footprint.

\section{Methods}
\label{sec:methods}

This section describes how we evaluate GFMs for AGB regression.

\subsection{Geospatial Foundation Models}
\label{sec:gfms}

We evaluate the 11 GFM backbones available on PANGAEA: CROMA \citep{croma}, DOFA \citep{dofa}, GFM-Swin \citep{gfmswin,han2024bridging}, Prithvi \citep{prithvi}, RemoteCLIP \citep{remoteclip}, SatlasNet \citep{satlas}, ScaleMAE \citep{scalemae}, SpectralGPT \citep{spectralgpt}, SSL4EO-MoCo \citep{wang2022ssl4eo}, TerraMind (tiny) \citep{terramind}, and Prithvi v2 (100M) \citep{prithvi2}. Of the multiple variants available for the SSL4EO model, we chose the MoCo variant, having the best reported score for the Finland biomass task out of all PANGAEA models \citep{pangaea2024}. We evaluate each GFM in two settings: with a frozen backbone, and with parameter-efficient fine-tuning via Low-Rank Adaptation (LoRA) \citep{hu2022lowrank}. We do not consider full end-to-end fine-tuning, as downstream practitioners typically lack the compute to fine-tune large backbones. The majority of these models (CROMA, DOFA, Prithvi, Prithvi v2, SatlasNet, SpectralGPT, SSL4EO-MoCo, and TerraMind) were pre-trained on multispectral EO data and can consume most of the Sentinel-2 band stack; while some also accommodate Sentinel-1 (C-band SAR). The remaining models (GFM-Swin, RemoteCLIP, ScaleMAE) were pre-trained on RGB imagery only, and can only consume the RGB Sentinel-2 bands. None of the benchmarked GFMs natively support the L-band SAR data used in the AGBD, despite its superiority over C-band SAR in characterizing vegetation properties \citep{Imhoff1995,Naidoo2015}. Some of the models (SatlasNet, Prithvi v1 and v2) support time-series. 

We also evaluate TESSERA and AEF embeddings, distributed as ready-to-use, per-pixel products, that we sample at the AGBD locations. 

The AEF embeddings \citep{alphaearth} form a global dataset of $64$-dimensional per-pixel features at 10m resolution, summarizing the multi-source EO observations acquired throughout each calendar year (optical, radar, thermal, elevation, climate). Importantly, AEF is trained to reconstruct a broad set of \emph{target} modalities (including GEDI LiDAR) from a smaller set of \emph{input} modalities (Sentinel-2, Sentinel-1, and Landsat-8/9). To prevent data leakage from GEDI pre-training targets, we exclude all AGBD validation and test samples that intersect with the published AEF pre-training locations from our analyses. We further supplement the AEF embeddings with a subset of AGBD modalities that were not reconstructed during pre-training (land-cover, slope, aspect, and sine--cosine encoded latitude/longitude), a setting we define as \textbf{AEF$^+$}. This tests whether the embeddings already encode simple ancillary information, or whether adding them yields complementary gains and additional robustness.

The TESSERA v1 embeddings \citep{feng2025tesseratemporalembeddingssurface} consist of $128$-dimensional per-pixel features at 10m resolution, summarizing the Sentinel-1 and Sentinel-2 observations acquired throughout each year. The embeddings carry no spatial context beyond the individual pixel. TESSERA releases its model weights alongside the embeddings, so users can either download the pre-computed product or generate the embeddings themselves.

\subsection{Predictive heads}
\label{sec:heads}

To regress biomass from the GFMs' features, we consider predictive heads of increasing capacity.

\paragraph{Linear Probe (LP)} A single linear layer that maps the 64-dimensional (for AEF), respectively 128-dimensional (for TESSERA) embeddings to a biomass estimate. This measures the linear separability of biomass information in the embedding space. Each model is trained three times with different random seeds, and we report the mean and standard deviation.

\paragraph{Multi Layer Perceptron (MLP)} A fully connected network with a single hidden layer of $256$ units and ReLU activation. This model captures non-linear relationships between embeddings and biomass. The hidden size is kept fixed at $256$ across both embedding products so that the head capacity is identical regardless of input dimensionality, isolating the effect of the representation from that of the head. Each model is trained three times with different random seeds, and we report the mean and standard deviation.

\paragraph{Fully convolutional network (\texttt{fcn\_film})} Our supervised baseline is a state-of-the-art (SOTA) model for biomass estimation, a fully-convolutional neural network of $\approx$\,2M parameters, adapted from prior work on canopy height estimation \citep{lang_canopy} and biomass estimation \citep{agbd2024}. It was shown in \citet{agbd2024} to surpass other models for AGB estimation. Our adapted version uses stochastically jittered Feature-wise Linear Modulation \citep{perez2018film} to perform implicit ensembling, as proposed by \citet{filmensemble}, to provide well-calibrated estimates of epistemic uncertainty with low computational overhead. 

\paragraph{UPerNet} The UPerNet head \citep{Xiao2018}, with Pyramid Pooling Module and a lateral connection Feature Pyramid Network, that the PANGAEA framework uses as its standard dense-prediction head. As per PANGAEA's configuration, the size of the head (in millions of parameters) varies depending on encoder size, as follows: $\approx$\,30M for the TerraMind tiny models, SSL4EO, SatlasNet, GFM-Swin, AEF and TESSERA embeddings; $\approx$\,40M for DOFA, Prithvi and Prithvi v2, RemoteCLIP, CROMA, ScaleMAE; and 165M for SpectralGPT.

Evaluating geospatial embeddings with lightweight LP and MLP heads is standard practice in the literature \citep{alphaearth,feng2025tesseratemporalembeddingssurface}. Their low computational cost also makes them our heads of choice for AGBD Full, whose size makes training heavier heads impractical. The \texttt{fcn\_film} model plays a dual role: trained end-to-end on the raw AGBD features, it is our fully supervised SOTA baseline, while on top of the embeddings it acts as a high-capacity head. As for the UPerNet head, it allows for a fair comparison of AEF and TESSERA embeddings with the other GFMs.

\begin{figure}[!ht]
    \centering
    \includegraphics[width=\textwidth]{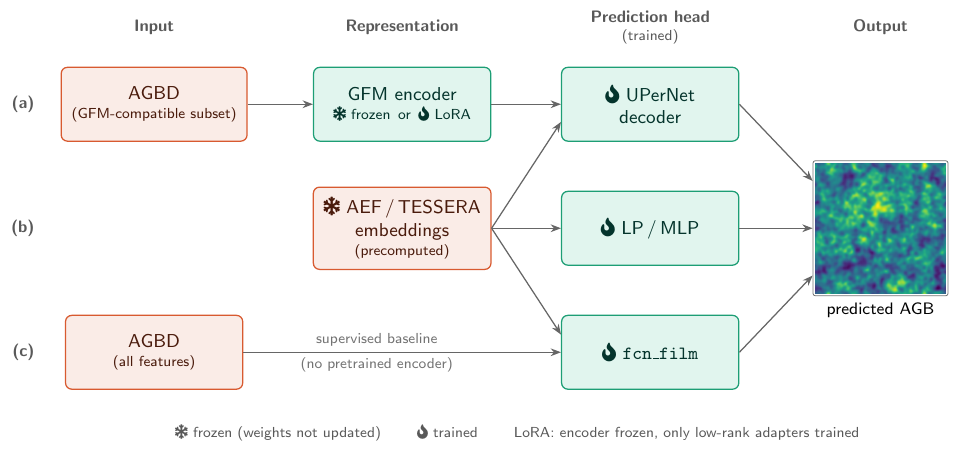}
    \caption{Biomass estimation framework. Each row is one way of obtaining an input representation at a GEDI footprint; columns read left to right, from input to predicted AGB. \textbf{(a)}~The subset of AGBD features compatible with the GFM's pre-training inputs is encoded by the GFM, kept either frozen or fine-tuned with LoRA, and decoded by a trained UPerNet head. \textbf{(b)}~The pre-computed AEF and TESSERA embedding products are used as fixed inputs to a UPerNet head, a lightweight linear probe (LP) or MLP head, or \texttt{fcn\_film}. \textbf{(c)}~As a supervised baseline, \texttt{fcn\_film} is trained on all AGBD features without any pre-trained encoder. Snowflakes (\faSnowflake) mark frozen components and flames (\faFire) trained ones; with Low-Rank Adaptation (LoRA), only low-rank adapters are added to the frozen encoder and trained.}
    \label{fig:pipeline}
\end{figure}

\section{Experiments}
\label{sec:experiments}

To limit computational overhead, we first screen all methods on AGBD Lite and train only the best-performing ones on AGBD Full. Beyond aggregate performance, a central requirement for any global mapping method is that it generalizes spatially and temporally beyond the training dataset. We therefore design two sets of experiments that stress-test generalization across space (Section \ref{sec:geo_gen}) and across time (Section \ref{sec:temp_gen}). Finally, we compare our predictions with those of an operational product, on independent reference data (Section \ref{sec:agbref_eval}). 

\subsection{Geographical Generalization}
\label{sec:geo_gen}

We partition the datasets by continental regions (North America, South America, Africa, Europe, North Asia\footnote{Note that due to the geographical coverage of the GEDI L4A data (between 51.6 degrees N and S) we do not have any samples in the North Asia region.}, South Asia, Australasia) and define three evaluation protocols for each target region $R \in$ \{Africa, South Asia, South America\}:

\begin{enumerate}
    \item \textbf{within-region}: Train \emph{only} on data from region $R$, evaluate on test data from $R$ (localized model).
    \item \textbf{cross-region}: Train on all data \emph{except} region $R$, evaluate on $R$ (transfer to a region without local labels).
    \item \textbf{general}: Train on \emph{all} data including $R$, evaluate on $R$ (monolithic global model, default behavior).
\end{enumerate}

This design disentangles within-distribution performance from the ability to transfer to new regions with possibly unique properties. We focus our evaluation on Africa, South Asia, and South America as these regions represent the most ecologically diverse and carbon-dense terrestrial ecosystems, yet they remain under-represented in standard machine learning benchmarks compared to the Global North. 

\subsection{Temporal Generalization}
\label{sec:temp_gen}

The AGBD contains GEDI footprints from both 2019 and 2020. We define three training configurations to assess the temporal generalization abilities of the models:

\begin{enumerate}
    \item \textbf{cross-year}: Train exclusively on 2019 training data, evaluate on test data 2020.
    \item \textbf{within-year}: Train exclusively on 2020 training data, evaluate on test data 2020.
    \item \textbf{all-years}: Train on both years, evaluate on 2020 test data.
\end{enumerate}

Comparing these configurations quantifies the degree to which models capture inter-annual variations.

\subsection{Independent validation on AGBref}
\label{sec:agbref_eval}

\begin{figure}[!ht]
    \centering
    \includegraphics[width=1\linewidth]{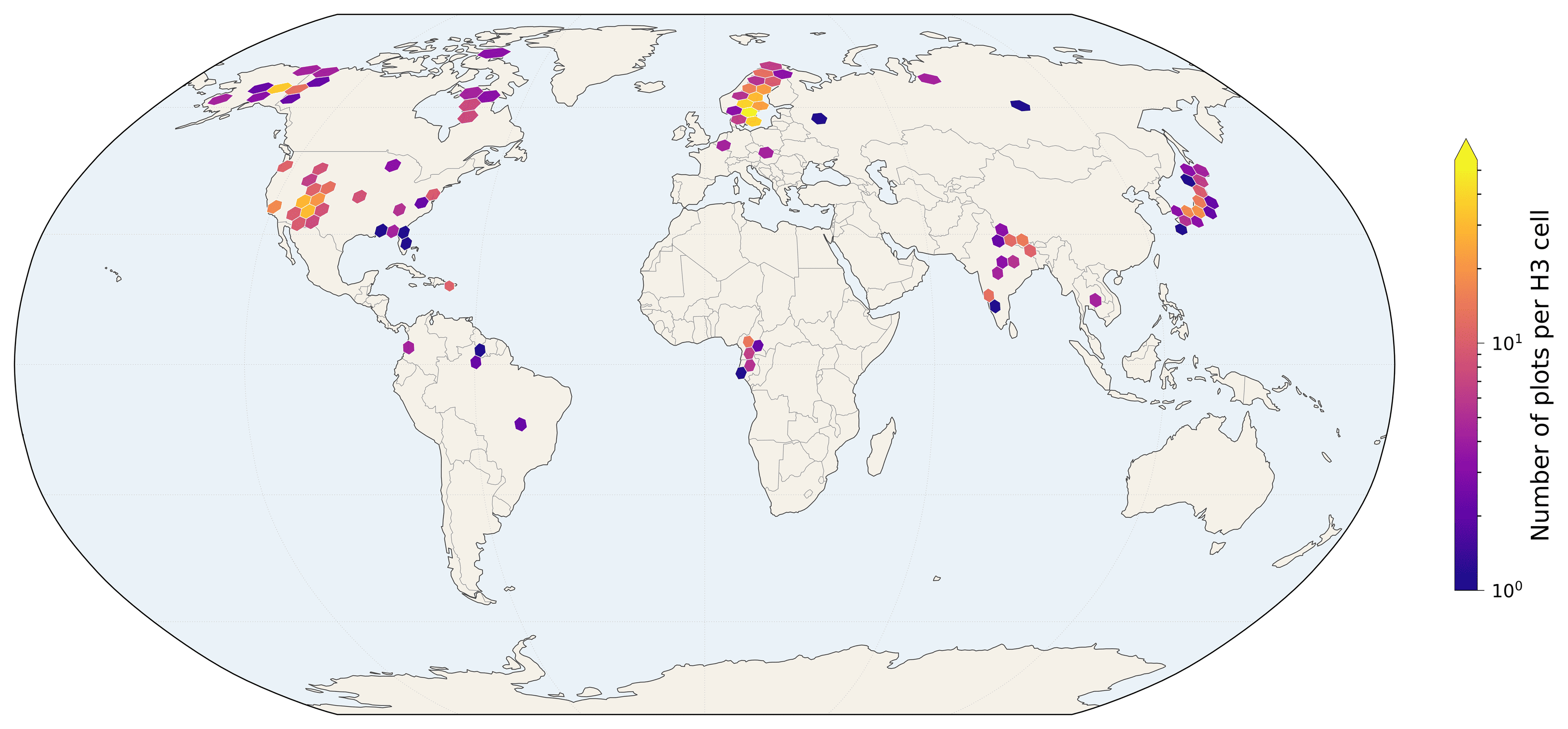}
    \caption{Spatial distribution of the AGBref plots used for independent validation (Robinson projection, H3 hexagonal grid with $\approx86$\si{km^2} per cell). Color: number of plots per cell (log scale).}
    \label{fig:agbref-density}
\end{figure}

To assess whether performance holds beyond the held-out AGBD test set, we compare our best model on AGBref \citep{AGBref}, a fully independent global reference dataset synthesized from a diverse array of national forest inventories, permanent research plots, and high-resolution airborne LiDAR maps. As a benchmark on the same plots, we also evaluate the operational ESA CCI Biomass map \citep{ESACCI}. AGBref provides multi-temporal biomass estimates across different resolutions, specifically 500m, 1km, 10km, and 25km. By providing this standardized framework, AGBref offers a necessary base for independent validation and inter-comparison of global biomass products. As a trade-off between uncertainty and resolution, we use the 10km version of the dataset, following the methods of \citet{SANTORO2026112536}. In Figure~\ref{fig:agbref-density}, we show the geographical distribution of the reference plots.

Specifically, we generate our predictions of AGB, and extract v6 ESA CCI Biomass maps, over the geographical and temporal extent of the AGBref dataset. As AGBref provides aggregated biomass values for $10\times10$~\si{km^2} regions, we aggregate the predicted maps to the same regions using the mean. To avoid spatial autocorrelation, we remove the $\approx3\%$ of plots that intersect with the AGBD training set. 

\section{Results}
\label{sec:results}

\subsection{Comparison of GFMs and supervised baselines on AGBD}
\label{sec:main_results}

\begin{figure}[!t]
    \centering
    \includegraphics[width=1\linewidth]{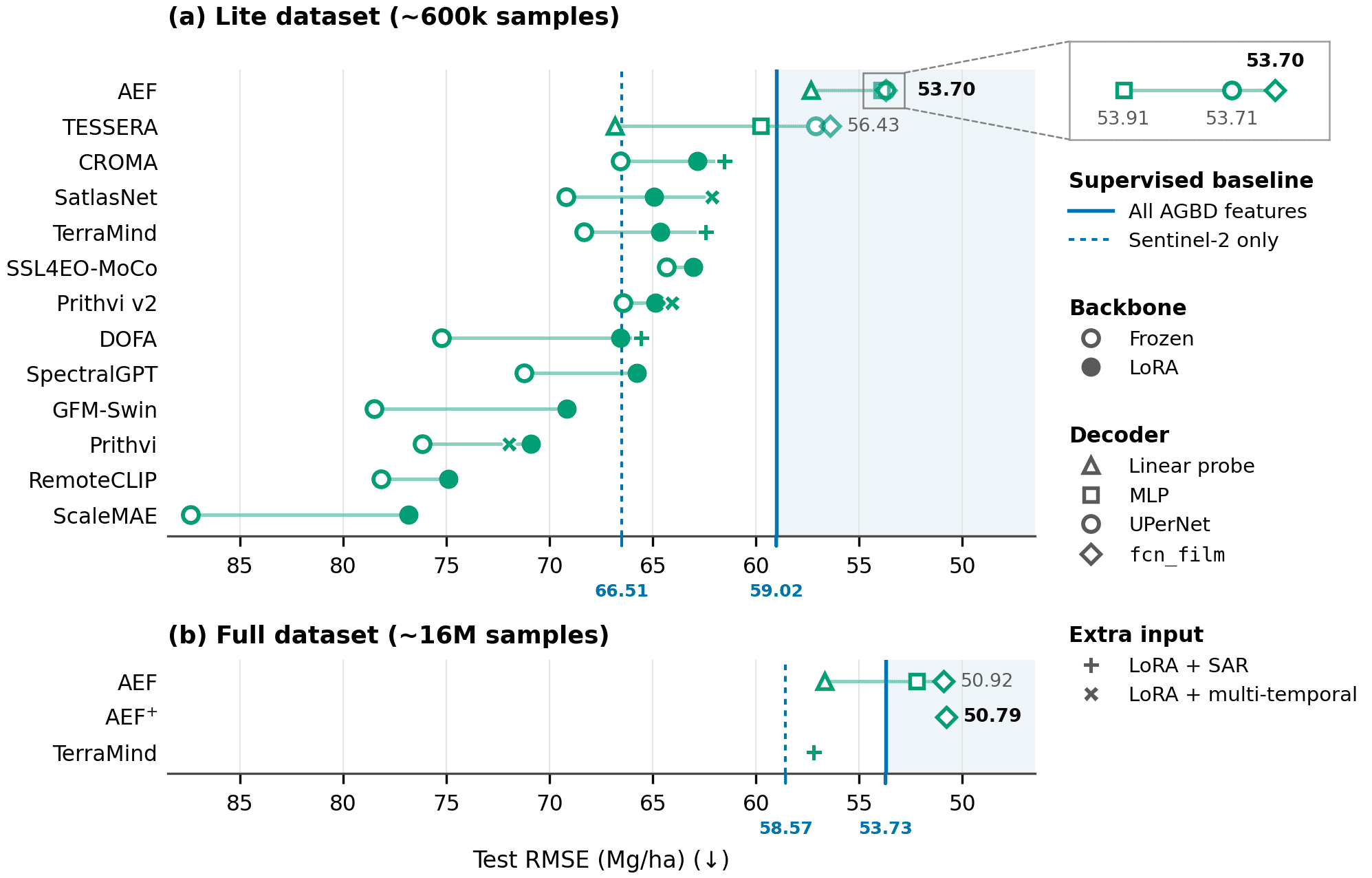}
    \caption{Test RMSE (Mg/ha) with inverted axis. In both panels, blue lines mark the supervised \texttt{fcn\_film} baseline trained from scratch on all AGBD features (solid) or on Sentinel-2-only (dashed); the shaded area marks results better than the baseline on all AGBD features. Exact numbers for all markers can be found in Table \ref{tab:all-results}. (a) Models trained and evaluated on AGBD Lite. All the GFMs use a UPerNet head on a frozen or LoRA-adapted backbone, optionally with L-band SAR ($+$) or multi-temporal Sentinel-2 ($\times$) input. Additionally, AEF and TESSERA embeddings are shown with heads of lower capacity: linear probe ($\triangle$), MLP ($\square$) and \texttt{fcn\_film} ($\lozenge$). (b) Models trained and evaluated on AGBD Full; AEF embeddings alone and with ancillary features (AEF$^+$: land cover, slope, aspect and encoded coordinates), with various prediction heads, and TerraMind (+SAR, LoRA, UPerNet head).}
    \label{fig:benchmark}
\end{figure}

Figure~\ref{fig:benchmark}a summarizes the test root mean squared error (RMSE) on AGBD Lite ($\approx600$k samples) for the 13 GFMs, all paired with a UPerNet head, alongside the two supervised \texttt{fcn\_film} baselines. The AEF and TESSERA embeddings were the two most accurate models overall ($53.70$ and $56.43$\mgha, respectively). They were also the only models to outperform the supervised baseline trained on all AGBD features ($59.02$\mgha). 

With all encoders frozen, AEF and TESSERA outperformed the other GFMs by a large margin, with 17--39\% and 11--35\% lower RMSE, respectively. Otherwise, only SSL4EO-MoCo ($64.34$\mgha) improved on the supervised baseline trained only on Sentinel-2 data ($66.51$\mgha), with CROMA and Prithvi v2 performing comparably. The remaining encoders were considerably worse. Fine-tuning with LoRA improved every encoder's performance, with the largest gain for the encoders whose frozen features performed worst. After fine-tuning, 6 of the 11 encoders outperformed the Sentinel-2-only baseline, but the gap to AEF and TESSERA remained (15--30\% and 9--26\% lower RMSE, respectively). 

For encoders that support additional inputs, we also fine-tuned with L-band SAR (ALOS PALSAR-2) for CROMA, TerraMind and DOFA, or with multi-temporal Sentinel-2 imagery for SatlasNet, Prithvi and Prithvi v2. These additions were overall beneficial for the models, although these effects remain smaller than the effect of fine-tuning itself. The best resulting configuration, CROMA (+SAR) ($61.56$\mgha) remained behind AEF and TESSERA, as well as the supervised baseline trained on multi-modal data ($13\%$, $7\%$ and $4\%$ lower RMSE, respectively). 

On AGBD Lite, we also paired the AEF and TESSERA embeddings with lighter heads (Fig.~\ref{fig:benchmark}a). With a per-pixel MLP, AEF reaches $53.91$\mgha, and even a linear probe on AEF ($57.34$\mgha) outperforms the supervised baseline ($59.02$\mgha). TESSERA is more sensitive to the choice of head: its MLP ($59.79$\mgha) falls just short of the supervised baseline, and its linear probe ($66.86$\mgha) is on par with the Sentinel-2-only baseline ($66.51$\mgha).

\begin{figure}[!t]
    \centering
    \includegraphics[width=0.9\linewidth]{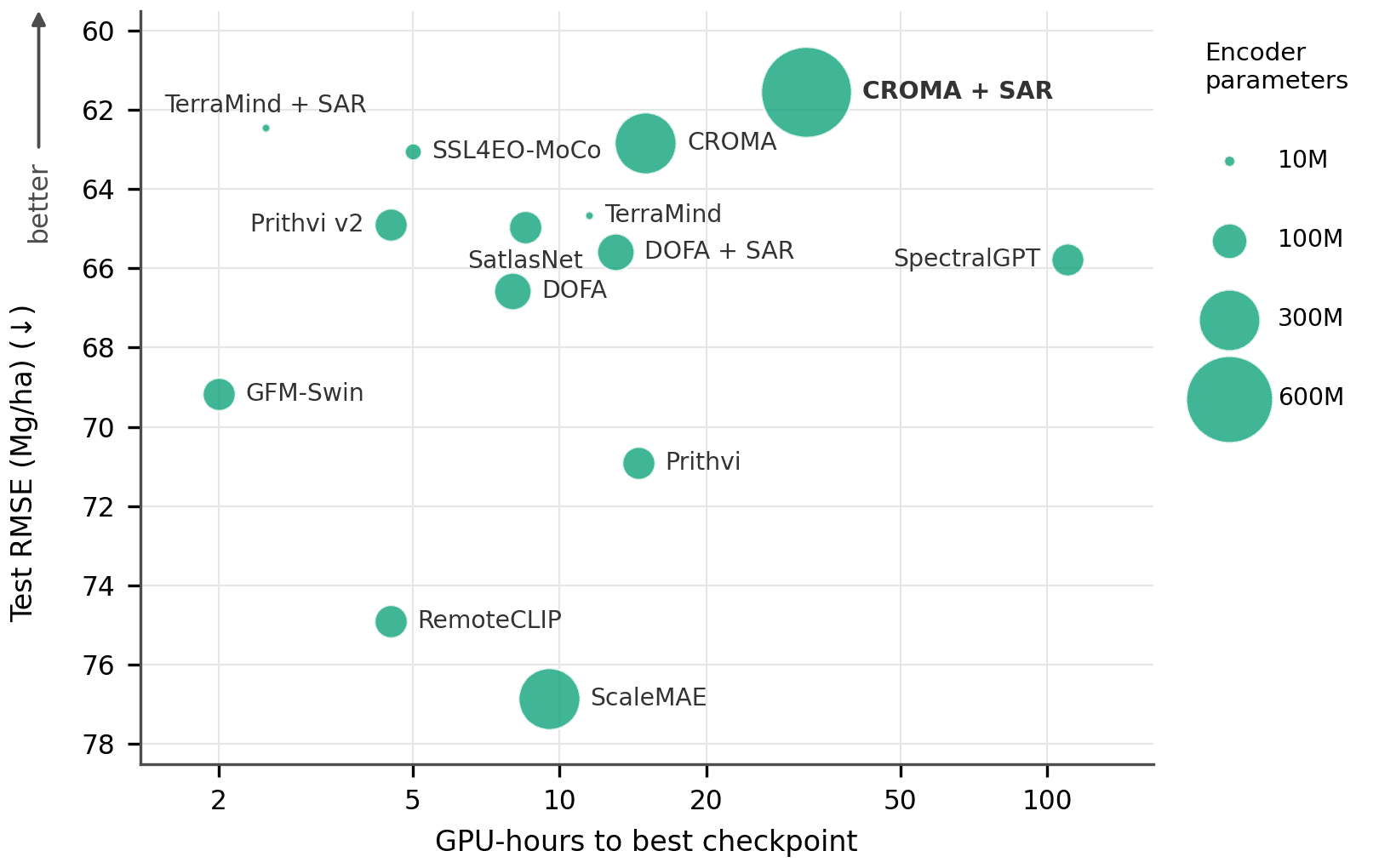}
    \caption{GPU-hours until the checkpoint with the lowest validation RMSE (log scale, NVIDIA GeForce RTX 4090) against test RMSE (inverted axis) for each GFM fine-tuned with LoRA and a UPerNet head on AGBD Lite; +SAR denotes added L-band SAR input. AEF and TESSERA are not shown because their embeddings are pre-computed. Marker area is proportional to the number of encoder parameters (excluding the head and LoRA adapters).}
    \label{fig:cost}
\end{figure}

Figure~\ref{fig:cost} relates accuracy to model size and fine-tuning cost, for the 11 encoders. AEF and TESSERA are not shown because their embeddings are pre-computed and only the head is trained. TerraMind (+SAR) used one of the smallest encoders and, within 2.5 GPU-hours of fine-tuning, reached $62.45$\mgha, within $1$\mgha of CROMA (+SAR), which used the largest encoder and roughly $32$ GPU-hours. 

\begin{figure}[!t]
    \centering
    \includegraphics[width=1\linewidth]{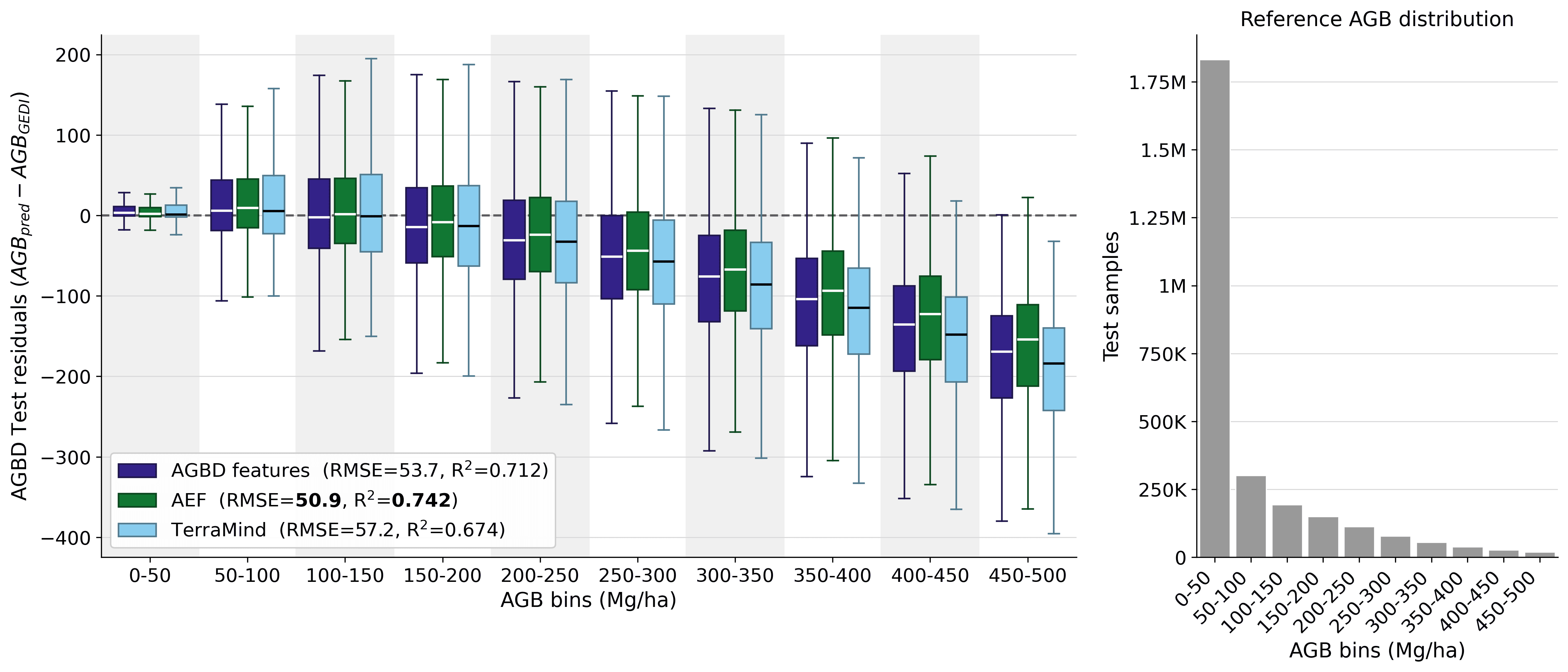}
    \caption{Test residuals (predicted $-$ reference AGB, Mg/ha) of \texttt{fcn\_film} trained on AGBD features (purple), on AEF embeddings (green), and of TerraMind (+SAR) (LoRA, UPerNet head; light blue), all trained on AGBD Full and evaluated on its test set. Left: residuals by $50$\mgha bin of reference AGB (boxes: median and interquartile range; whiskers: most extreme values within 1.5$\times$IQR of the box; outliers not shown); the legend gives each model’s overall RMSE and $R^2$.  Right: number of test samples per reference AGB bin.}
    \label{fig:binned_residuals}
\end{figure}

Building on these results on the AGBD Lite, we selected AEF for experiments on AGBD Full. We trained AEF-based models on AGBD Full ($\approx16$M samples) with prediction heads of varying capacity (Fig.~\ref{fig:benchmark}b). A linear probe on the AEF embeddings ($56.67$\mgha) outperformed the Sentinel-2-only baseline, but not the multi-modal one, while an MLP head ($52.22$\mgha) surpassed both. An \texttt{fcn\_film} head achieved $50.92$\mgha, surpassing the baseline. Adding ancillary features to the embeddings (AEF$^+$) marginally reduced the RMSE further, to $50.79$\mgha. To assess whether additional training data narrows the gap between fine-tuned encoders and AEF, we also trained TerraMind (+SAR) on AGBD Full. We selected TerraMind rather than CROMA (+SAR) because they reached similar accuracy while the former required about an order of magnitude less compute (Fig.~\ref{fig:cost}). TerraMind (+SAR) reached $57.21$\mgha, still trailing behind AEF. 

Figure~\ref{fig:binned_residuals} shows test residuals stratified into $50$\mgha bins of reference AGB for three models: AEF with an \texttt{fcn\_film} head, the supervised baseline trained on AGBD features, and the TerraMind (+SAR) model. The reference distribution is strongly long-tailed: about two thirds of test samples lie below $50$\mgha and fewer than $2\%$ exceed $400$\mgha. Beyond $200$\mgha, all models increasingly underestimate biomass, with AEF being the least affected.

\subsection{Geographical generalization}
\label{sec:geo_results}

Table~\ref{tab:geo_africa} presents results across three continents under the \textit{cross-region}, \textit{within-region}, and \textit{general} protocols. We compare \texttt{fcn\_film} models trained on AGBD features, AEF embeddings, and AEF$^+$ features under various configurations.

\begin{table}[h]
\centering
\caption{Test RMSE (Mg/ha) ($\downarrow$) of \texttt{fcn\_film} trained on AGBD Full with AGBD features, AEF embeddings or AEF$^+$ (AEF plus land cover, slope, aspect and encoded coordinates) for geographical generalization to three target regions. \textbf{cross-region}: trained without the target region; \textbf{within-region}: trained only on the target region; \textbf{general}: trained on all regions including target. All models are evaluated on the target region’s test samples. Mean $\pm$ std over three runs. Best results per region and protocol in \textbf{bold}, second best results \underline{underlined}.}
\label{tab:geo_africa}
\begin{tabular}{llccc}
\toprule
\textbf{Region} & \textbf{Features} & \textbf{cross-region} & \textbf{within-region} & \textbf{general} \\
\midrule
\multirow{3}{*}{Africa}
& AGBD & 40.79 {\footnotesize $\pm$ 0.10} & 31.42 {\footnotesize $\pm$ 0.01} & 31.47 {\footnotesize $\pm$ 0.05} \\
& AEF  & \underline{35.08} {\footnotesize $\pm$ 0.09} & \underline{29.17} {\footnotesize $\pm$ 0.01} & \hspace{-3pt}\textbf{29.21} {\footnotesize $\pm$ 0.02} \\
& AEF$^+$ & \textbf{33.72} {\footnotesize $\pm$ 0.12} & \textbf{29.13} {\footnotesize $\pm$ 0.01} & \underline{29.24} {\footnotesize $\pm$ 0.02} \\
\midrule
\multirow{3}{*}{South Asia}
& AGBD & \textbf{79.28} {\footnotesize $\pm$ 0.07} & 77.17 {\footnotesize $\pm$ 0.04} & 73.65 {\footnotesize $\pm$ 0.11} \\
& AEF  & 99.34 {\footnotesize $\pm$ 0.31} & \hspace{-3pt}\textbf{69.44} {\footnotesize $\pm$ 0.01} & \hspace{-3pt}\textbf{69.75} {\footnotesize $\pm$ 0.04} \\
& AEF$^+$ & \underline{80.85} {\footnotesize $\pm$ 0.23} & \underline{69.50} {\footnotesize $\pm$ 0.01} & \underline{70.49} {\footnotesize $\pm$ 0.08} \\
\midrule
\multirow{3}{*}{South America}
& AGBD & 42.40 {\footnotesize $\pm$ 0.04} & 28.85 {\footnotesize $\pm$ 0.02} & 27.77 {\footnotesize $\pm$ 0.04} \\
& AEF  & \underline{32.05} {\footnotesize $\pm$ 0.08} & \underline{25.62} {\footnotesize $\pm$ 0.03} & \textbf{25.61} {\footnotesize $\pm$ 0.01} \\
& AEF$^+$ & \textbf{32.02} {\footnotesize $\pm$ 0.08} & \textbf{25.56} {\footnotesize $\pm$ 0.01} & \underline{25.62} {\footnotesize $\pm$ 0.01} \\
\bottomrule
\end{tabular}
\end{table}

\begin{figure}[!t]
    \centering
    \includegraphics[width=1.0\linewidth]{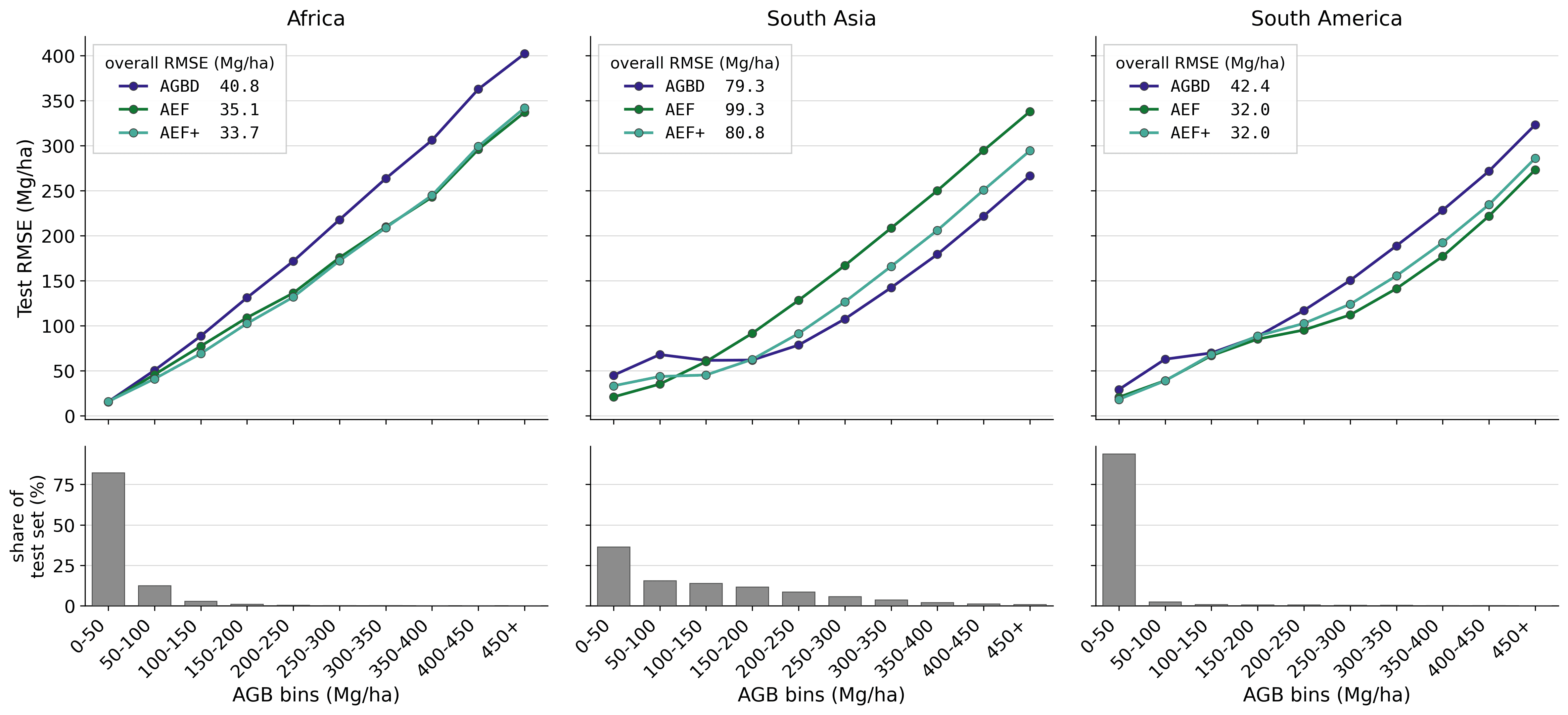}
    \caption{Test RMSE by AGB bin under the \textit{cross-region} protocol, for \texttt{fcn\_film} trained on AGBD features (purple), AEF (green) and AEF$^+$ (AEF plus land cover, slope, aspect and encoded coordinates) (light blue). Each column is one target region, scored by a model trained without any data from that region. Top row: RMSE within 50\mgha bins of reference AGB; the legend gives each model's overall RMSE (Mg/ha). Bottom row: share of that region's test footprints in each bin.}
    \label{fig:sa_binned}
\end{figure}

Across all three regions and training protocols, AEF and AEF$^+$ embeddings consistently outperform raw AGBD features. When in-region training data is available (\textit{within-region} and \textit{general}), AEF and AEF$^+$ reduce RMSE by 2--3\mgha in Africa and South America, and by 4--8\mgha in South Asia. AEF and AEF$^+$ perform comparably under these protocols, with neither consistently dominating the other. The most substantial gains emerge under \textit{cross-region} transfer, where no in-region data is available during training. In Africa, AEF$^+$ reduces RMSE from 40.79 to 33.72\mgha (17.3\%), and in South America, AEF and AEF$^+$ reduce RMSE from 42.40 to approximately 32\mgha (24.5\%). However, this transfer is not uniform across regions. In South Asia, AEF under \textit{cross-region} produces an RMSE of 99.34\mgha, 25\% worse than AGBD features (79.28\mgha), while AEF$^+$ (80.85\mgha) performs on par with AGBD features. Stratifying the error by AGB (Fig.~\ref{fig:sa_binned}) shows that this deficit is confined to high biomass.

\subsection{Temporal generalization}
\label{sec:temp_results}

Table~\ref{tab:temporal} compares models trained on 2019--2020 (all years), 2019 (cross-year) and 2020 (within-year) data, always evaluated on the 2020 test set. We compare \texttt{fcn\_film} models trained on AGBD features, AEF embeddings, and AEF$^+$ features under these three configurations.
 
\begin{table}[h]
\centering
\caption{AGBD test RMSE (Mg/ha) ($\downarrow$) on the 2020 AGBD test samples for \texttt{fcn\_film} trained on AGBD features, AEF embeddings or AEF$^+$ models, using training data from all years (2019 and 2020), cross-year (2019 only) or within-year (2020 only). Mean $\pm$ std over three runs. Best results per column in \textbf{bold}, second best results \underline{underlined}. Test tiles are spatially disjoint from the training tiles in all configurations.}
\label{tab:temporal}
\begin{tabular}{lccc}
\toprule
& \multicolumn{3}{c}{\textbf{Trained on}} \\
\textbf{Features} & \textbf{all years} & \textbf{cross-year} & \textbf{within-year} \\
\midrule
AGBD & 53.29 {\footnotesize $\pm$ 0.02} & 59.12 {\footnotesize $\pm$ 0.05} & 52.20 {\footnotesize $\pm$ 0.01} \\
AEF  & \underline{50.64} {\footnotesize $\pm$ 0.03} & \hspace{-4pt}\textbf{52.23} {\footnotesize $\pm$ 0.01} & \underline{51.05} {\footnotesize $\pm$ 0.01} \\
AEF$^+$ & \hspace{-4pt}\textbf{50.52} {\footnotesize $\pm$ 0.01} & \underline{52.52} {\footnotesize $\pm$ 0.02} & \hspace{-4pt}\textbf{50.92} {\footnotesize $\pm$ 0.01} \\
\bottomrule
\end{tabular}
\end{table}

AEF and AEF$^+$ consistently outperform AGBD features across all three protocols, though the margins are more modest than in the geographical generalization experiments. When training data is temporally matched to the evaluation year (within-year), the gap between AGBD features and AEF$^+$ narrows ($52.20$ vs.\ $50.92$\mgha). When a one-year gap is introduced (cross-year), AGBD features degrade to $59.12$\mgha while AEF and AEF$^+$ remain at $52.23$ and $52.52$\mgha, respectively. Notably, AEF and AEF$^+$ trained on cross-year data alone still outperform AGBD features trained on all years ($53.29$\mgha). The all-years protocol offers only marginal further improvement over cross-year for AEF and AEF$^+$.

\subsection{Qualitative comparison of prediction maps}
\label{sec:qualitative}

\begin{figure}[p]
    \centering
    \includegraphics[width=1.0\linewidth]{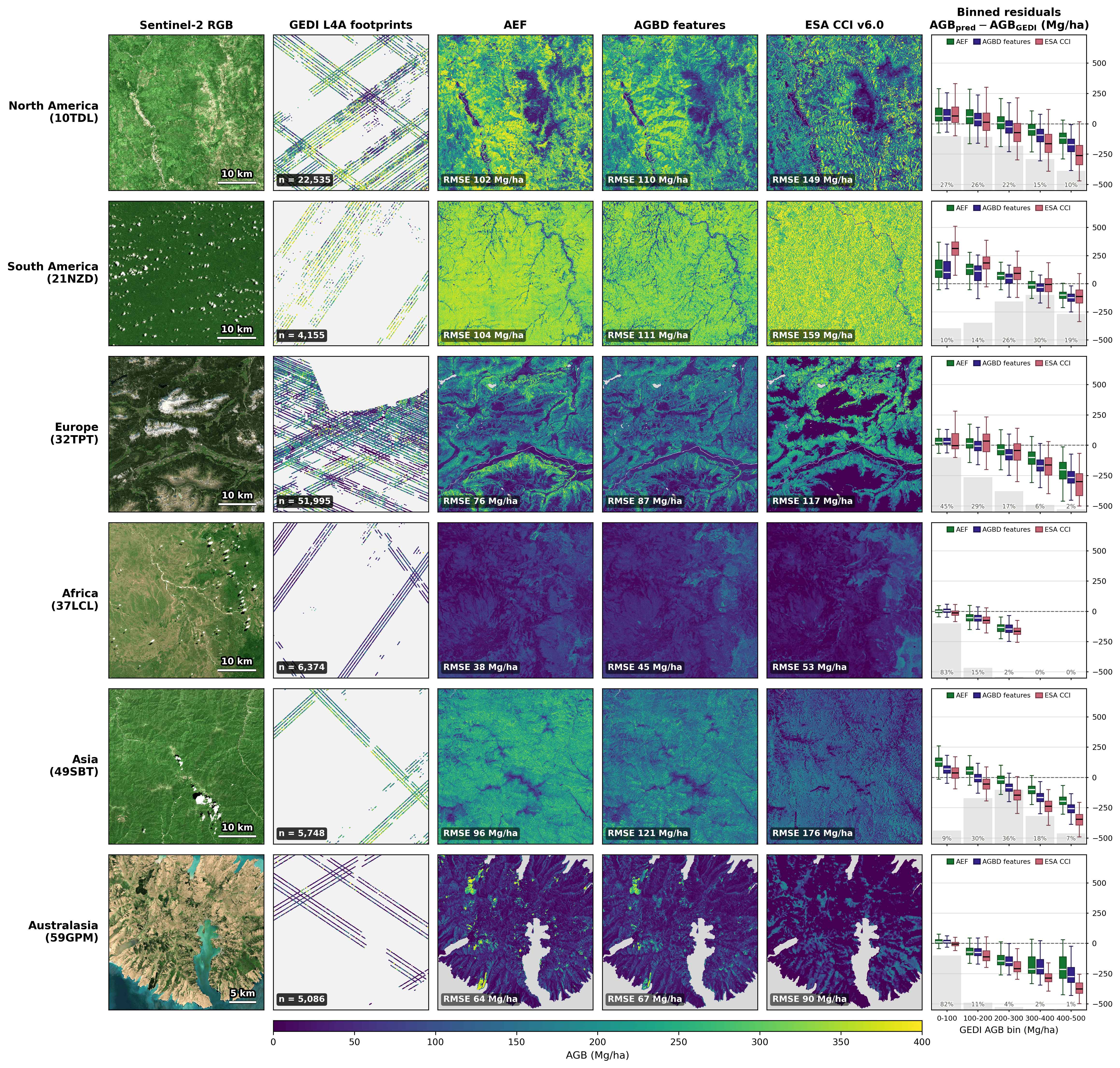}
    \caption{Qualitative comparison of predicted AGB maps across three sources on six zoomed-in windows. Rows, top to bottom: North America (10TDL, Six Rivers National Forest, northern California, USA), South America (21NZD, Maripasoula, French Guiana), Europe (32TPT, Tyrol, Austria), Africa (37LCL, Selous Game Reserve, Lindi Region, southern Tanzania), Asia (49SBT, Qinling, Shaanxi, China) and Australasia (tile 59GPM, Banks Peninsula, New Zealand). Columns, left to right: Sentinel-2 true-color composite; GEDI L4A footprints; \texttt{fcn\_film} on AEF embeddings; \texttt{fcn\_film} on AGBD features (supervised baseline); the ESA CCI v6.0 (100m) biomass product; residuals (predicted - GEDI, Mg/ha) per GEDI AGB bin. The RMSE (Mg/ha) in each panel is computed against the available GEDI L4A footprints inside the window. Color scale: AGB (Mg/ha).}
    \label{fig:pred_maps}
\end{figure}

The results so far are aggregate error metrics. To complement them, we inspect the spatial behavior of the different methods directly, on zoomed-in windows of six Sentinel-2 tiles spanning distinct biomes and hemispheres: North America (10TDL, Six Rivers National Forest, northern California, USA), South America (21NZD, Maripasoula, French Guiana), Europe (32TPT, Tyrol, Austria), Africa (37LCL, Selous Game Reserve, Lindi Region, southern Tanzania), Asia (49SBT, Qinling, Shaanxi, China) and Australasia (tile 59GPM, Banks Peninsula, New Zealand). For each tile we generate dense wall-to-wall AGB maps with \texttt{fcn\_film} trained on AEF embeddings (our best GFM) and on AGBD features (our supervised baseline), and we extract the co-located ESA CCI v6.0 product. We additionally report the RMSE of every panel against the independent GEDI L4A footprints that fall inside the displayed window, alongside binned residuals. Figure~\ref{fig:pred_maps} shows the resulting maps. The per-tile GEDI RMSE mirrors the aggregate benchmark ranking of Section~\ref{sec:main_results}: on all six tiles the AEF-based model attains the lowest error ($102$, $104$, $76$, $38$, $96$ and $64$\mgha), ahead of the supervised AGBD-features baseline ($110$, $111$, $87$, $45$, $121$ and $67$\mgha) and, furthest from the GEDI reference, ESA CCI ($149$, $159$, $117$, $53$, $176$ and $90$\mgha). Visually, the AEF maps recover spatially coherent biomass gradients that follow terrain and vegetation structure while staying smooth over homogeneous cover, the AGBD-features baseline reproduces the same large-scale patterns but carries more small-scale variability, and ESA CCI (100m) is the smoothest of all, diverging most from the GEDI footprints.

\subsection{Comparison to the ESA CCI Biomass map on AGBref}
\label{sec:agbref}

\begin{figure}[!t]
    \centering
    \includegraphics[width=0.85\linewidth]{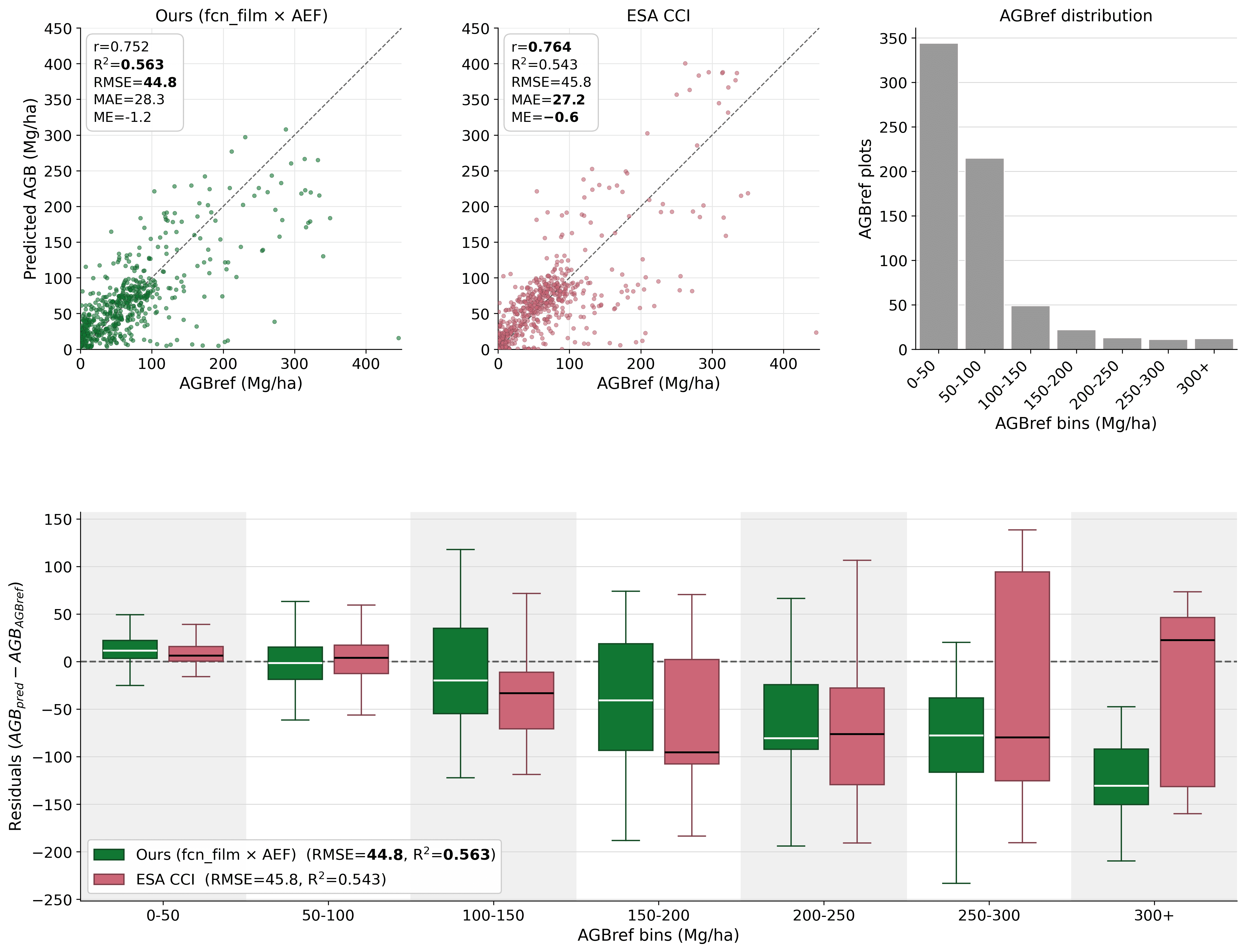}
    \caption{Comparison of predicted AGB against AGBref reference values aggregated to 10km cells, for \texttt{fcn\_film} on AEF embeddings (green) and ESA CCI v6.0 (red) on the \textit{All} configuration. Top row: per-plot scatter plots with the 1:1 line (dashed), and the distribution of AGBref values (right). Bottom row: residuals (predicted $-$ reference, Mg/ha) per AGBref bin; boxes show the median and interquartile range, whiskers extend to the most extreme residual within 1.5$\times$IQR of the box; the legend gives overall RMSE and $R^2$.}
    \label{fig:agbref}
\end{figure}

We evaluate our model using two distinct configurations: \textit{All}, which includes all plots outside of Japan; and \textit{Subset}, which is restricted to plots within a 1500km radius of an AGBD sample ($\approx78\%$ of the \textit{All} set). The \textit{Subset} configuration facilitates a more equitable comparison; while the ESA CCI maps were trained on a global scale, our model was trained specifically on the AGBD regions. Overall results are reported in Table \ref{tab:model_comparison}, and binned performance plots are shown in Figure~\ref{fig:agbref}.

\begin{table}[!t]
    \centering
        \caption{Evaluation of \texttt{fcn\_film} on AEF embeddings and of ESA CCI v6.0 against AGBref reference plots, after aggregating both maps to AGBref’s 10km cells. \textit{All} includes all plots except Japan, and \textit{Subset} is restricted to plots within 1500km of an AGBD sample. $R^2$: coefficient of determination; $r$: Pearson correlation coefficient; RMSE: Root Mean Squared Error; MAE: Mean Absolute Error; ME: Mean Error. RMSE, MAE and ME in Mg/ha. Best values in bold.}
    \begin{tabular}{p{2cm}lccccc}\toprule
        \textbf{Plots} & \multicolumn{1}{c}{\textbf{Model}} & \textbf{$R^2$} & \textbf{$r$} & \textbf{RMSE} & \textbf{MAE} & \textbf{ME} \\ \midrule
        
        \multirow{2}{=}{All} & \texttt{fcn\_film} $\times$ AEF & \textbf{0.563}& 0.752 & \textbf{44.79}& 28.31 & -1.18\\ 
        & ESA CCI & 0.543 & \textbf{0.764}& 45.77 & \textbf{27.20}& \textbf{-0.60} \\
        
        \addlinespace[6pt] 
        
        \multirow{2}{=}{Subset} & \texttt{fcn\_film} $\times$ AEF & \textbf{0.696}& 0.837 & \textbf{34.32}& 24.18 & \textbf{4.42} \\ 
        & ESA CCI & 0.687 & \textbf{0.862}& 34.83 & \textbf{22.13}& 4.72\\ \bottomrule
    \end{tabular}

    \label{tab:model_comparison}
\end{table}

Across both configurations, our \texttt{fcn\_film} model trained on AEF and the ESA CCI maps achieve near-parity in $R^2$, $r$ and RMSE. Noticeable, albeit small, differences emerge in MAE, where ESA CCI holds a slight advantage in both configurations, and in ME (bias), where the two models trade places: ESA CCI has the smaller absolute bias on the \textit{All} set ($-0.60$ vs.\ $-1.18$\mgha) and our model on the \textit{Subset} ($4.42$ vs.\ $4.72$\mgha), with all biases remaining small. Performance improves from the \textit{All} to \textit{Subset} configuration for both models; our model's improvement is expected given the proximity to its training regions, while the CCI improvement may reflect either inherent predictability of these regions or overlap with CCI's own reference data.
Recall that all comparisons are performed after aggregating to AGBref's 10km resolution, meaning that short-range biomass variations visible in the two EO-based map products are suppressed, no matter whether they are in agreement or not. Both models show a characteristic saturation effect, underestimating AGB at high reference values (Fig.~\ref{fig:agbref}, top row), but track the 1:1 line well at low-to-moderate biomass ($0$--$100$\mgha; Fig.~\ref{fig:agbref}, bottom row). In the $100$--$250$\mgha range, our model (\texttt{fcn\_film}$\times$AEF) maintains medians closer to zero and smaller spread than ESA CCI, demonstrating greater stability against signal saturation. Above $300$\mgha, our model exhibits strong, consistent underestimation while ESA CCI shows an apparent shift toward positive medians, but this behavior is largely driven by high variance and sparse sample counts in those bins (Fig.~\ref{fig:agbref}, top right).

\section{Discussion}
\label{sec:discussion}

We find large performance differences between GFMs for above-ground biomass estimation. Across frozen and fine-tuned settings, only 2 of the 13 evaluated models outperform the state-of-the-art supervised baseline. The other 11 remain below it even after LoRA fine-tuning, in line with \citet{muszynski2024}, who found fine-tuned GFMs to perform slightly below a supervised baseline trained from scratch for biomass estimation in Brazil. Several factors contribute to this gap, which we reflect on below. We then turn to how these embeddings generalize across space and time, and how they stand against an operational biomass product.

\subsection{What makes a GFM useful for biomass regression}

The largest difference among the GFMs is the observational context compressed into each representation. AEF and TESSERA summarize a full year of observations per pixel, from Sentinel-1 and Sentinel-2 for TESSERA, and from a broader set of modalities for AEF (e.g.\ thermal, elevation). Most of the other GFMs encode a single acquisition, mostly from optical bands. For biomass estimation, this is a limiting factor: optical reflectance captures it only indirectly and saturates over closed canopies, whereas radar backscatter and the seasonal trajectory of a pixel can carry complementary information on canopy structure \citep{pauls2025capturing}. Adding radar backscatter during fine-tuning only slightly improves the performance of the GFMs supporting it (although these models natively accommodate C-band SAR, and we provide L-band SAR). Adding temporal context, likewise, gives the relevant GFMs modest and inconsistent gains (significant for SatlasNet, very small for Prithvi v2, and negative for Prithvi). Beyond their inputs, the pre-training objectives of most of these GFMs may also play a role: masked image modeling and contrastive learning yield patch-token representations optimized for reconstruction or instance discrimination, rather than for regressing continuous variables such as biomass.

LoRA fine-tuning improved the performance of every GFM it was applied to, with the biggest gains for the models whose frozen features performed worst. This suggests that frozen performance partly reflects a mismatch between the AGBD inputs and each model's pre-training conditions, rather than the content of its representation. We identify two sources of mismatch, which are not specific to biomass and likely affect other PANGAEA datasets as well. First, each GFM is pre-trained with different (and hardly documented) Sentinel-2 processing strategies, so there might be a distribution shift. Second, most of the GFMs are pre-trained to expect fixed $224\times224$ pixel inputs, so the small AGBD patch must either be upsampled, presenting textures at a scale the encoders never saw during pre-training, or padded, which preserves their scale but leaves the actual observations to make up little of the input ($62.45$\mgha upsampled vs. $65.41$\mgha padded, for LoRA-adapted TerraMind +SAR). To some extent, fine-tuning absorbs these mismatches and improves on the frozen features performance, but the models still under-perform with respect to AEF and TESSERA. More recent models show promising developments in this regard: THOR \citep{forgaard2026thor} is designed to be deployable with any patch size, SMARTIES \citep{Sumbul2025} is sensor-agnostic, and UniverSat \citep{perron2026universat} and RAMEN \citep{Houdre_2026_CVPR} are resolution- and modality-agnostic.

Furthermore, AEF's edge over TESSERA seems to be due to two factors. First, TESSERA encodes each pixel's time series independently, so its embeddings carry no spatial context. With a per-pixel MLP head, TESSERA does not improve on the supervised \texttt{fcn\_film} trained on the single-date AGBD features, despite encoding a full year of Sentinel-1/2 observations; only with larger spatial heads does it beat the baseline (56.43\mgha with \texttt{fcn\_film} and 57.12\mgha with UPerNet). The AEF performance, on the other hand, barely degrades when the spatial head is removed (53.70 vs.\ 53.91\mgha). Second, AEF is trained as a self-supervised autoencoder whose dense, per-pixel reconstruction \emph{targets} include GEDI relative height metrics. TESSERA, by contrast, learns only from its own Sentinel-1/2 inputs and is never exposed to structural signals of this kind. 

In summary, we hypothesize that it is the combination of spatio-temporal context, multi-modal inputs, aligned objective, and pre-computed delivery format that gives AEF its edge over the other GFMs and the state-of-the-art supervised baseline. Beyond performance, the practical implications are significant. AEF embeddings are pre-computed, eliminating the need for GPU-intensive feature extraction. A simple MLP trained on these embeddings outperforms the state-of-the-art \texttt{fcn\_film} model trained on remote sensing features, at a fraction of the computational and data engineering costs. For resource-constrained settings, this efficiency advantage is considerable. 
  
\subsection{Behavior beyond the training distribution}
The geographical generalization experiments reveal that AEF embeddings not only improve average performance but also change how models behave when training and evaluation distributions diverge. Under \textit{cross-region} transfer, AEF and AEF$^+$ reduce RMSE by 14--25\% in Africa and South America. Only the biomass head is trained without target-region data: the AEF encoder was pre-trained on globally distributed data, so the embeddings carry a global prior that raw features lack. Across all three regions, AEF and AEF$^+$ also maintain stable performance between the \textit{within-region} and \textit{general} protocols, indicating that pooling data from diverse regions neither helps nor hinders the embeddings, unlike AGBD features which show more variable behavior under multi-region training. However, this transfer is not universally reliable: in South Asia, AEF trained without any local data produces an RMSE 25\% worse than raw AGBD features. Other studies also report mixed cross-region transfer with AEF embeddings, for canopy height \citep{hamoudzadeh2026}, agricultural tasks \citep{ma2026harvesting} and irrigated cropland mapping \citep{yang2026irrigated}; the latter attribute the decline to region-specific associations learned during supervised training rather than to the embeddings alone. These results point to complementary roles for the two embedding strategies. When training data is representative of the target region, AEF and AEF$^+$ perform similarly, and the simpler AEF formulation suffices. When representative data cannot be guaranteed (as is likely in operational global-scale deployment where training coverage will inevitably be uneven) AEF$^+$ offers a more robust alternative, maintaining competitive performance with AEF in the general setting, and with AGBD features in the cross-region setting. In both cases, learned embeddings stabilize behavior across training protocols, effectively closing the gap between localized and general models. This is especially relevant since globally distributed external reference data for evaluation is scarce, meaning such failures could easily go undetected.
 
The temporal generalization results indicate that AEF and AEF$^+$ encode more temporally invariant representations than raw spectral features. When a one-year gap is introduced between training and evaluation data, AGBD features degrade sharply while AEF and AEF$^+$ remain largely stable. This asymmetry suggests that the learned embeddings abstract away from year-specific phenological or atmospheric conditions to which raw spectral features are sensitive. The practical consequence is that AEF and AEF$^+$ trained on cross-year data alone still outperform AGBD features trained on all years, indicating that the embeddings can compensate for missing temporal coverage; a relevant property for operational deployment, where satellite imagery is available almost in real-time, whereas the provision of training labels may lag behind.
 
Finally, our evaluation against AGBref demonstrates that a model that leverages AEF embeddings, trained on a subset of globally distributed regions, can achieve near-parity with the globally calibrated ESA CCI biomass product. In the controlled-subset comparison, both models achieve essentially identical $R^2$ and RMSE, while our model exhibits smaller absolute bias (ME). This result is notable given that the CCI maps are produced using globally distributed reference data and sophisticated multi-sensor fusion, whereas our model was trained on only the 11 regions from the AGBD, suggesting that the quality of the learned representation can partially compensate for limited geographic coverage in the training data.

\subsection{Implications for practitioners}

The decisive property of a GFM for biomass estimation is the information its representation encodes. When that information is rich enough, a lightweight head suffices: trained on AEF embeddings, an MLP outperforms the supervised state-of-the-art model trained on the AGBD, at a much lower cost. Pre-computed embeddings, however, come with constraints. Their producer fixes the input sensors, the temporal aggregation and the spatial resolution. When the model weights are not released, as is the case for AEF, the embeddings cannot be fine-tuned nor generated for other time periods, and their continued availability depends on a single provider. TESSERA, the second best-performing GFM in our benchmark, releases both its weights and its embeddings, offering an open alternative for practitioners wary of AEF's proprietary model.

\subsection{Limitations}
\label{sec:limitations}
Several limitations should be considered when interpreting our results. First, our training labels come from GEDI L4A, itself a model-derived product with known uncertainties \citep{Pascual2023,Jia2023,Li2024}. Our benchmark therefore measures the ability to predict GEDI-derived AGB, and a model that reproduces GEDI's systematic biases would be rewarded for doing so. The evaluation against AGBref partially mitigates this: AGBref is derived from forest inventories, research plots and airborne LiDAR, so its error structure is largely independent of GEDI L4A's, and agreement between our predictions and AGBref is evidence that the models capture biomass rather than GEDI-specific artifacts. Second, the AEF embeddings, while publicly available as a dataset, were produced by a proprietary model whose architecture and training data are not fully disclosed. This limits reproducibility of the embedding generation step, though downstream experiments using the released embeddings remain fully reproducible. It also restricts users to the years released by the provider, and to an annual temporal resolution. Furthermore, our temporal generalization experiments span only two consecutive years (2019--2020); longer-term generalization, including sensitivity to disturbance events and land-use change, remains untested. We also could not identify why AGBD features transfer better than AEF to South Asia’s high-biomass forests: land cover, terrain, the shift of the embeddings from the training data and the height-to-biomass relation of the GEDI L4A labels did not explain it. Finally, while it is informative of expected global performance, our comparison against AGBref reference data is by definition limited to the geographical extent of the considered plots.

\section{Conclusion}
\label{sec:conclusion}
We benchmarked 13 geospatial foundation models for global-scale above-ground biomass regression on the AGBD, using their frozen representations and, when applicable, LoRA fine-tuning. All were compared against a fully supervised baseline, and the best-performing GFM was further evaluated for geographical and temporal generalization and benchmarked against an operational biomass product on independent reference data.

The GFMs showcase very different performance for this task. Eleven of them did not beat the supervised baseline trained on multi-modal data ($59.02$\mgha), whether frozen ($64.34$\mgha for SSL4EO-MoCo) or fine-tuned ($61.56$\mgha for CROMA +SAR), and at a far higher computational cost. Two of them (AEF and TESSERA) outperform that baseline ($53.71$\mgha and $57.12$\mgha, respectively). This gap is primarily attributable to the restricted input support of some of these models: most of them are confined to a single time step, and to limited input modalities. Multi-temporal support yields modest gains, and the models pre-trained on C-band SAR moderately benefit from L-band fine-tuning; but not enough to reach TESSERA and AEF levels. Architectural rigidity compounds this, as most models expect a fixed patch size and force the input to be resized or padded to match, while the absence of documented pre-processing pipelines leaves the risk of distribution shift at inference difficult to rule out. Fine-tuning seems to absorb this mismatch, but cannot supply the missing information. On the other hand, AEF and TESSERA distill a full year of multi-modal observations into a compact per-pixel representation. 

AEF's advantage over the supervised baseline is largest when labels are scarce (AGBD Lite), but persists on AGBD Full. A single-hidden-layer MLP trained on AEF embeddings ($52.22$\mgha) already outperforms the supervised baseline ($53.73$\mgha) and the fine-tuned TerraMind (+SAR) model ($57.21$\mgha). Increasing the capacity of the head (\texttt{fcn\_film}) further improves performance ($50.92$\mgha) and augmenting the embeddings with a few covariates gives the best overall result ($50.79$\mgha). AEF also confers stronger geographical and temporal generalization than raw EO features, though not unconditionally (e.g., in South Asia). Finally, our \texttt{fcn\_film} model trained on 11 regions with AEF embeddings performs on par with the globally calibrated ESA CCI product when both are evaluated against the independent AGBref plots, suggesting that strong global representations acquired during pre-training can partly compensate for narrow geographical training coverage.

Taken together, our results indicate that GFM features can be highly informative for above-ground biomass regression when the underlying model is trained on rich multi-modal, multi-temporal data, especially when they are served to users as an analysis-ready embedding layer.

\section*{Data availability}

The code and data supporting this study are publicly available at
\url{https://github.com/ghjuliasialelli/AGBD-GFMs}.

\section*{CRediT authorship contribution statement}

\textbf{Ghjulia Sialelli:} Conceptualization, Methodology, Software,
Writing -- original draft, Writing -- review \& editing.
\textbf{Linus Scheibenreif:} Supervision, Writing -- review \& editing.
\textbf{Jan Dirk Wegner:} Supervision, Writing -- review \& editing.
\textbf{Konrad Schindler:} Funding acquisition, Supervision,
Writing -- review \& editing.

\section*{Declaration of generative AI and AI-assisted technologies in the writing process}
During the preparation of this work, the authors used Anthropic's Claude to assist with restructuring the manuscript and with drafting and language-editing portions of the introduction, related work, and discussion. The tool was not used to generate, analyze, or interpret any data, or experimental results. After using this tool, the authors reviewed and edited the content as needed and take full responsibility for the content of the publication.

\section*{Acknowledgments}
GS is supported by an ETH AI Center doctoral fellowship. We thank Valerio Marsocci for his support in our integration of AGBD Lite and AGBD Full in the PANGAEA framework. We thank Frank Feng for generating the TESSERA embeddings for our regions and years of interest. The AlphaEarth Foundations Satellite Embedding dataset is produced by Google and Google DeepMind. We express our gratitude to Taylor Geospatial for providing the AEF embeddings free of charge on Source Cooperative.

\clearpage

\appendix

\section{Main results}

Table~\ref{tab:all-results} gives the exact RMSE values behind every marker and reference line in Figures~\ref{fig:benchmark} and~\ref{fig:cost}. 

\newcolumntype{C}[1]{>{\centering\arraybackslash}p{#1}}
\newsavebox{\tabA}
\newsavebox{\tabB}
\newsavebox{\tabC}
\newlength{\rightw}
\newsavebox{\leftcol}
\newlength{\leftH}

\begin{table}[h]
\centering
\small
\sbox{\tabA}{%
\begin{tabular}{lcc}\toprule
     \textbf{Model} & \multicolumn{2}{c}{\textbf{RMSE ($\downarrow$)}}\\\midrule
                & \textbf{frozen} & \textbf{LoRA}\\ \cmidrule(lr){2-3}
     SSL4EO-MoCo     & 64.34 & 63.05\\
     Prithvi    & 76.17 & 70.91\\
     \quad \textit{+MT}        & —     & 71.98\\
     Prithvi v2 & 66.43 & 64.89\\
     \quad \textit{+MT}        & —     & 64.08\\
     CROMA      & 66.57 & 62.83\\
     \quad \textit{+SAR}       & —     & \textbf{61.56}\\
     TerraMind  & 68.33 & 64.66\\
     \quad \textit{+SAR}       & —     & 62.45\\
     SatlasNet  & 69.20 & 64.96\\
     \quad \textit{+MT}        & —     & \underline{62.14}\\
     SpectralGPT& 71.23 & 65.77\\
     DOFA       & 75.24 & 66.57\\
     \quad \textit{+SAR}       & —     & 65.58\\
     RemoteCLIP & 78.17 & 74.91\\
     GFM-Swin   & 78.52 & 69.17\\
     ScaleMAE   & 87.39 & 76.85\\
     AEF        & \textbf{53.71} & —\\
     TESSERA    & \underline{57.12} & —\\ \bottomrule
\end{tabular}}%
\sbox{\tabB}{%
\begin{tabular}{lccc}\toprule
     \textbf{Features} & \multicolumn{3}{c}{\textbf{RMSE ($\downarrow$)}}\\\midrule
             & \textbf{LP} & \textbf{MLP} & \textbf{\texttt{fcn\_film}} \\ \cmidrule(lr){2-4}
     S2      & — & — & 66.51 {\footnotesize $\pm$ 0.04} \\
     AGBD    & — & — & 59.02 {\footnotesize $\pm$ 0.05} \\
     AEF     & 57.34 {\footnotesize $\pm$ 0.01} & \underline{53.91} {\footnotesize $\pm$ 0.02} & \textbf{53.70} {\footnotesize $\pm$ 0.04} \\
     TESSERA & 66.86 {\footnotesize $\pm$ 0.06} & 59.79 {\footnotesize $\pm$ 0.09} & 56.43 {\footnotesize $\pm$ 0.02} \\ \bottomrule
\end{tabular}}%
\sbox{\tabC}{%
\begin{tabular*}{\wd\tabB}{@{\extracolsep{\fill}}lcc}
\toprule
\textbf{  Model} & \textbf{Features} & \textbf{RMSE ($\downarrow$)} \\
\midrule
\addlinespace[2pt]
\texttt{fcn\_film} & S2    & 58.57 {\footnotesize $\pm$ 0.03} \\
\texttt{fcn\_film} & AGBD    & 53.73 {\footnotesize $\pm$ 0.02} \\
LP                 & AEF\phantom{$^+$}     & 56.67 {\footnotesize $\pm$ 0.01} \\
MLP                & AEF\phantom{$^+$}     & 52.22 {\footnotesize $\pm$ 0.02} \\
\texttt{fcn\_film} & AEF\phantom{$^+$}     & \underline{50.92} {\footnotesize $\pm$ 0.02} \\
\texttt{fcn\_film} & AEF$^+$ & \hspace{-4pt}\textbf{50.79} {\footnotesize $\pm$ 0.01} \\
TerraMind & S2 + SAR & 57.21 \phantom{\footnotesize $\pm$ 0.00} \\
\addlinespace[2pt]
\bottomrule
\end{tabular*}}%
\setlength{\rightw}{\wd\tabB}%
\ifdim\wd\tabC>\rightw \setlength{\rightw}{\wd\tabC}\fi
\sbox{\leftcol}{%
\begin{minipage}[t]{\wd\tabA}
    \vspace{0pt}\centering
    \subcaption{GFMs on Lite.}\label{tab:gfm-lite}
    \usebox{\tabA}
\end{minipage}}%
\setlength{\leftH}{\dimexpr\ht\leftcol+\dp\leftcol\relax}%
\begin{adjustbox}{max width=\linewidth}
\usebox{\leftcol}%
\hspace{1.5em}%
\begin{minipage}[t][\leftH][t]{\rightw}
    \vspace{0pt}\centering
    \subcaption{Features and heads on Lite.}\label{tab:features-lite}
    \usebox{\tabB}

    \vfill
    \subcaption{Main configurations, AGBD Full.}\label{tab:main-results}
    \usebox{\tabC}
\end{minipage}
\end{adjustbox}

\caption{Test RMSE (Mg/ha, $\downarrow$). (a) GFMs trained and evaluated on AGBD Lite with a frozen or LoRA fine-tuned backbone, and UPerNet head; +SAR: with L-band ALOS PALSAR-2 input; +MT: with three Sentinel-2 dates (AGBD-MT Lite). (b) Sentinel-2 bands only (S2), all AGBD features (AGBD), and AEF and TESSERA embeddings with linear probe (LP), MLP or \texttt{fcn\_film} heads, trained and evaluated on AGBD Lite. (c) Main configurations on AGBD Full; AEF$^+$: AEF plus land cover, slope, aspect and encoded coordinates; TerraMind is LoRA-adapted with a UPerNet head. In (b) and (c), results are mean $\pm$ std over three runs, except for TerraMind; (a) reports single runs. Best results in bold, second best underlined.}
\label{tab:all-results}
\end{table}

\begin{figure}[!ht]
    \centering
    \begin{minipage}[c]{0.60\textwidth}
        \centering
        \includegraphics[width=\linewidth]{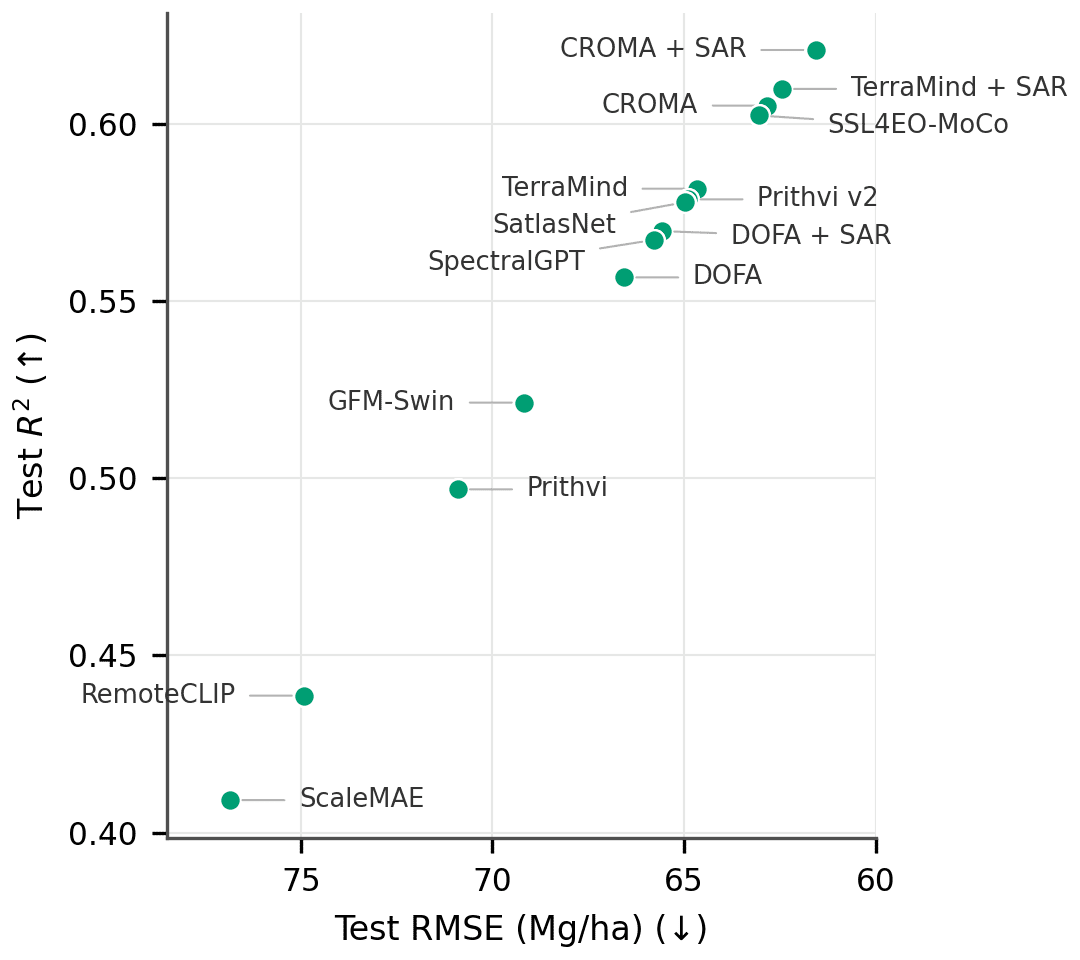} 
    \end{minipage}\hfill
    \begin{minipage}[c]{0.40\textwidth}
        \centering
        \scriptsize
        \begin{tabular}{lcc}
            \toprule
            Model & RMSE ($\downarrow$) & $R^2$ ($\uparrow$) \\
            \midrule
            SSL4EO-MoCo        & 63.05 & 0.602 \\
            Prithvi       & 70.91 & 0.497 \\
            Prithvi v2    & 64.89 & 0.579 \\
            CROMA         & 62.83 & 0.605 \\
            \quad +SAR    & \textbf{61.56} & \textbf{0.621} \\
            TerraMind     & 64.66 & 0.582 \\
            \quad +SAR    & \underline{62.45} & \underline{0.610} \\
            SatlasNet     & 64.96 & 0.578 \\
            SpectralGPT   & 65.77 & 0.567 \\
            DOFA          & 66.57 & 0.557 \\
            \quad +SAR    & 65.58 & 0.570 \\
            RemoteCLIP    & 74.91 & 0.439 \\
            GFM-Swin      & 69.17 & 0.521 \\
            ScaleMAE      & 76.85 & 0.409 \\
            \bottomrule
        \end{tabular}
    \end{minipage}
    \caption{Test RMSE (Mg/ha) against test $R^2$ for each GFM fine-tuned with LoRA and a UPerNet head on AGBD Lite; +SAR denotes added L-band SAR input. Left: the RMSE axis is inverted, so better models lie towards the upper right. Right: exact RMSE and $R^2$ values. Both metrics rank the models identically.}
    \label{fig:r2}
\end{figure}

Figure~\ref{fig:r2} compares the test RMSE and $R^2$ for the GFMs fine-tuned with LoRA on AGBD Lite. Both metrics give the same ranking. This agreement is expected: all models are evaluated on the same test set, for which $R^2 = 1 - \mathrm{RMSE}^2/\mathrm{Var}(y)$ is a strictly decreasing function of RMSE. We therefore report RMSE in the main text, as it is expressed in the units of the target variable (Mg/ha).

\clearpage
\bibliographystyle{elsarticle-harv}
\bibliography{main}

@String(CVPR  = {IEEE Conf. Comput. Vis. Pattern Recog.})

@String(ICCV  = {Int. Conf. Comput. Vis.})

@String(ECCV  = {Eur. Conf. Comput. Vis.})

@String(AAAI  = {AAAI})

@String(CVPR  = {CVPR})

@String(ICCV  = {ICCV})

@String(ECCV  = {ECCV})

@inproceedings{filmensemble,
 author = {Turkoglu, Mehmet Ozgur and Becker, Alexander and G\"{u}nd\"{u}z, H\"{u}seyin Anil and Rezaei, Mina and Bischl, Bernd and Daudt, Rodrigo Caye and D\textquotesingle Aronco, Stefano and Wegner, Jan and Schindler, Konrad},
 booktitle = {Advances in Neural Information Processing Systems},
 editor = {S. Koyejo and S. Mohamed and A. Agarwal and D. Belgrave and K. Cho and A. Oh},
 pages = {22229--22242},
 publisher = {Curran Associates, Inc.},
 title = {{FiLM-Ensemble}: Probabilistic Deep Learning via Feature-wise Linear Modulation},
 url = {https://proceedings.neurips.cc/paper_files/paper/2022/file/8bd31288ad8e9a31d519fdeede7ee47d-Paper-Conference.pdf},
 volume = {35},
 year = {2022}
}

@inproceedings{han2024bridging,
title={Bridging Remote Sensors with Multisensor Geospatial Foundation Models},
author={Boran Han and Shuai Zhang and Xingjian Shi and Markus Reichstein},
booktitle={Proceedings of the IEEE/CVF Conference on Computer Vision and Pattern Recognition (CVPR)},
pages={27852--27862},
year={2024},
doi={10.1109/CVPR52733.2024.02631}
}

@article{agbd2024,
  author = {Sialelli, Ghjulia and Peters, Torben and Wegner, Jan D. and Schindler, Konrad},
  title = {{AGBD}: A Global-scale Biomass Dataset},
  journal = {ISPRS Annals of the Photogrammetry, Remote Sensing and Spatial Information Sciences},
  volume = {X-G-2025},
  pages = {829-838},
  year = {2025},
  doi = {10.5194/isprs-annals-x-g-2025-829-2025},
  publisher = {Copernicus GmbH},
}

@article{pangaea2024,
      title = {{PANGAEA}: Assessing Geospatial Foundation Models Capabilities through a Global and Inclusive Benchmark},
      author={Valerio Marsocci and Yuru Jia and Georges {Le Bellier} and David Kerekes and Liang Zeng and Sebastian Hafner and Sebastian Gerard and Eric Brune and Ritu Yadav and Ali Shibli and Heng Fang and Yifang Ban and Maarten Vergauwen and Nicolas Audebert and Andrea Nascetti},
      journal = {IEEE Geoscience and Remote Sensing Magazine},
        volume = {14},
        number = {1},
        pages = {245--285},
        year = {2026},
        doi = {10.1109/MGRS.2025.3628194}
}

@article{prithvi,
  author = {Jakubik, Johannes and Roy, Sujit and Phillips, C. E. and Fraccaro, Paolo and Godwin, Denys and Zadrozny, Bianca and Szwarcman, Daniela and Gomes, Carlos and Nyirjesy, Gabby and Edwards, Blair and Kimura, Daiki and Simumba, Naomi and Chu, Linsong and Mukkavilli, S. Karthik and Lambhate, Devyani and Das, Kamal and Bangalore, Ranjini and Oliveira, Dario and Muszynski, Michal and Ankur, Kumar and Ramasubramanian, Muthukumaran and Gurung, Iksha and Khallaghi, Sam and Li, Hanxi and Cecil, Michael and Ahmadi, Maryam and Kordi, Fatemeh and Alemohammad, Hamed and Maskey, Manil and Ganti, Raghu and Weldemariam, Kommy and Ramachandran, Rahul},
  title = {Foundation Models for Generalist Geospatial Artificial Intelligence},
  journal = {arXiv preprint arXiv:2310.18660},
  year = {2023}
}

@inproceedings{scalemae,
  author = {Reed, Colorado J. and Gupta, Ritwik and Li, Shufan and Brockman, Sarah and Funk, Christopher and Clipp, Brian and Keutzer, Kurt and Candido, Salvatore and Uyttendaele, Matt and Darrell, Trevor},
  title = {{Scale-MAE}: A Scale-Aware Masked Autoencoder for Multiscale Geospatial Representation Learning},
  booktitle = {Proceedings of the IEEE/CVF International Conference on Computer Vision (ICCV)},
pages = {4065--4076},
year = {2023},
doi = {10.1109/ICCV51070.2023.00378}
}

@inproceedings{satlas,
  author = {Bastani, Favyen and Wolters, Piper and Gupta, Ritwik and Ferdinando, Joe and Kembhavi, Aniruddha},
  title = {{SatlasPretrain}: A Large-Scale Dataset for Remote Sensing Image Understanding},
  booktitle = {Proceedings of the IEEE/CVF International Conference on Computer Vision (ICCV)},
pages = {16726--16736},
year = {2023},
doi = {10.1109/ICCV51070.2023.01538}
}

@article{dofa,
  author = {Xiong, Zhitong and Wang, Yi and Zhang, Fahong and Stewart, Adam J. and Hanna, Joëlle and Borth, Damian and Papoutsis, Ioannis and {Le Saux}, Bertrand and Camps-Valls, Gustau and Zhu, Xiao Xiang},
  title = {Neural Plasticity-Inspired Multimodal Foundation Model for {Earth} Observation},
  journal = {arXiv preprint arXiv:2403.15356},
  year = {2024}
}

@inproceedings{croma,
  author = {Fuller, Anthony and Millard, Koreen and Green, James R.},
  title = {{CROMA}: Remote Sensing Representations with Contrastive Radar-Optical Masked Autoencoders},
  booktitle = {Advances in Neural Information Processing Systems},
volume = {36},
pages = {5506--5538},
year = {2023},
doi = {10.52202/075280-0241}
}

@article{remoteclip,
  author = {Liu, Fan and Chen, Delong and Guan, Zhangqingyun and Zhou, Xiaocong and Zhu, Jiale and Ye, Qiaolin and Fu, Liyong and Zhou, Jun},
  title = {{RemoteCLIP}: A Vision Language Foundation Model for Remote Sensing},
  journal = {IEEE Transactions on Geoscience and Remote Sensing},
volume = {62},
pages = {1--16},
year = {2024},
doi = {10.1109/TGRS.2024.3390838}
}

@article{spectralgpt,
  author = {Hong, Danfeng and Zhang, Bing and Li, Xuyang and Li, Yuxuan and Li, Chenyu and Yao, Jing and Yokoya, Naoto and Li, Hao and Ghamisi, Pedram and Jia, Xiuping and Plaza, Antonio and Gamba, Paolo and Benediktsson, Jon Atli and Chanussot, Jocelyn},
  title = {{SpectralGPT}: Spectral Remote Sensing Foundation Model},
  journal = {IEEE Transactions on Pattern Analysis and Machine Intelligence},
volume = {46},
number = {8},
pages = {5227--5244},
year = {2024},
doi = {10.1109/TPAMI.2024.3362475}
}

@misc{alphaearth,
  doi = {10.48550/ARXIV.2507.22291},
  url = {https://arxiv.org/abs/2507.22291},
  author = {Brown,  Christopher F. and Kazmierski,  Michal R. and Pasquarella,  Valerie J. and Rucklidge,  William J. and Samsikova,  Masha and Zhang,  Chenhui and Shelhamer,  Evan and Lahera,  Estefania and Wiles,  Olivia and Ilyushchenko,  Simon and Gorelick,  Noel and Zhang,  Lihui Lydia and Alj,  Sophia and Schechter,  Emily and Askay,  Sean and Guinan,  Oliver and Moore,  Rebecca and Boukouvalas,  Alexis and Kohli,  Pushmeet},
  title = {{AlphaEarth} Foundations: An embedding field model for accurate and efficient global mapping from sparse label data},
  publisher = {arXiv},
  year = {2025},
  copyright = {Creative Commons Attribution 4.0 International}
}

@article{lang_canopy,
  author = {Lang, Nico and Jetz, Walter and Schindler, Konrad and Wegner, Jan D.},
  title = {A High-Resolution Canopy Height Model of the {Earth}},
  journal = {Nature Ecology \& Evolution},
  volume = {7},
  pages = {1778--1789},
  year = {2023}
}

@inproceedings{gfmswin,
  author = {Mendieta, Mat{\'\i}as and Han, Boran and Shi, Xingjian and Zhu, Yi and Chen, Chen},
  title = {Towards Geospatial Foundation Models via Continual Pretraining},
  booktitle = {Proceedings of the IEEE/CVF International Conference on Computer Vision (ICCV)},
pages = {16760--16770},
year = {2023},
doi = {10.1109/ICCV51070.2023.01541}
}

@inproceedings{terramind,
  title={{TerraMind}: Large-Scale Generative Multimodality for {Earth} Observation},
  author={Jakubik, Johannes and Yang, Felix and Blumenstiel, Benedikt and Scheurer, Erik and Sedona, Rocco and Maurogiovanni, Stefano and Bosmans, Jente and Dionelis, Nikolaos and Marsocci, Valerio and Kopp, Niklas and Ramachandran, Rahul and Fraccaro, Paolo and Brunschwiler, Thomas and Cavallaro, Gabriele and Bernabe-Moreno, Juan and Longépé, Nicolas},
  booktitle={Proceedings of the IEEE/CVF International Conference on Computer Vision (ICCV)},
pages={7383--7394},
year={2025},
doi={10.1109/ICCV51701.2025.00693}
}

@article{prithvi2,
  author = {Szwarcman, Daniela and Roy, Sujit and Fraccaro, Paolo and Gíslason, {\TH}orsteinn Elí and Blumenstiel, Benedikt and Ghosal, Rinki and de Oliveira, Pedro Henrique and de Sousa Almeida, Joao Lucas and Sedona, Rocco and Kang, Yanghui and Chakraborty, Srija and Wang, Sizhe and Gomes, Carlos and Kumar, Ankur and Gaur, Vishal and Truong, Myscon and Godwin, Denys and Khallaghi, Sam and Lee, Hyunho and Hsu, Chia-Yu and Asanjan, Ata Akbari and Mujeci, Besart and Shidham, Disha and Balogun, Rufai Omowunmi and Kolluru, Venkatesh and Keenan, Trevor and Arevalo, Paulo and Li, Wenwen and Alemohammad, Hamed and Olofsson, Pontus and Mayer, Timothy and Hain, Christopher and Kennedy, Robert and Zadrozny, Bianca and Bell, David and Cavallaro, Gabriele and Watson, Campbell and Maskey, Manil and Ramachandran, Rahul and Bernabe-Moreno, Juan},
  title = {{Prithvi-EO-2.0}: A Versatile Multitemporal Foundation Model for {Earth} Observation Applications},
  journal = {IEEE Transactions on Geoscience and Remote Sensing},
  volume = {64},
  pages = {1--20},
  year = {2026},
  doi = {10.1109/TGRS.2025.3642610}
}

@inproceedings{feng2025tesseratemporalembeddingssurface,
  title={{TESSERA}: Temporal embeddings of surface spectra for {Earth} representation and analysis},
  author={Feng, Zhengpeng and Atzberger, Clement and Jaffer, Sadiq and Knezevic, Jovana and Sormunen, Silja and Young, Robin and Lisaius, Madeline C and Immitzer, Markus and Jackson, Toby and Ball, James and Coomes, David A. and Madhavapeddy, Anil and Blake, Andrew and Keshav, Srinivasan},
  booktitle={Proceedings of the IEEE/CVF Conference on Computer Vision and Pattern Recognition},
  pages={34818--34831},
  year={2026}
}

@inproceedings{nascetti2023biomassters,
title={{BioMassters}: A Benchmark Dataset for Forest Biomass Estimation using Multi-modal Satellite Time-series},
author={Andrea Nascetti and Ritu Yadav and Kirill Brodt and Qixun Qu and Hongwei Fan and Yuri Shendryk and Isha Shah and Christine Chung},
booktitle={Advances in Neural Information Processing Systems (Datasets and Benchmarks Track)},
volume={36},
pages={20409--20420},
year={2023},
doi={10.52202/075280-0896}
}

@inproceedings{gordon2026mmearthbench,
title={{MMEarth-Bench}: Global Model Adaptation via Multimodal Test-Time Training},
author={Lucia Gordon and Serge Belongie and Christian Igel and Nico Lang},
booktitle={Computer Vision -- ECCV 2026},
series={Lecture Notes in Computer Science},
publisher={Springer},
pages={334--353},
year={2026},
doi={10.1007/978-3-032-37252-9_19}
}

@inproceedings{wang2025unifiedcopernicusfoundationmodel,
      title={Towards a Unified {Copernicus} Foundation Model for {Earth} Vision}, 
      author={Yi Wang and Zhitong Xiong and Chenying Liu and Adam J. Stewart and Thomas Dujardin and Nikolaos Ioannis Bountos and Angelos Zavras and Franziska Gerken and Ioannis Papoutsis and Laura Leal-Taixé and Xiao Xiang Zhu},
      booktitle={Proceedings of the IEEE/CVF International Conference on Computer Vision (ICCV)},
      pages={9888--9899},
      year={2025},
      doi={10.1109/ICCV51701.2025.00922}
}

@inproceedings{fibaek2024PhilEO,
    title={{PhilEO Bench}: {Evaluating Geo-Spatial Foundation Models}},
    author={Fibaek, Casper and Camilleri, Luke and Luyts, Andreas and Dionelis, Nikolaos and {Le Saux}, Bertrand},
    year={2024},
    booktitle={Proceedings of the IEEE International Geoscience and Remote Sensing Symposium (IGARSS)},
    pages={2739--2744},
    doi={10.1109/IGARSS53475.2024.10642780}
  }

@article{bountos2023fomo,
  title={{FoMo}: Multi-Modal, Multi-Scale and Multi-Task Remote Sensing Foundation Models for Forest Monitoring},
  author={Bountos, Nikolaos Ioannis and Ouaknine, Arthur and Papoutsis, Ioannis and Rolnick, David},
  journal={Proceedings of the AAAI Conference on Artificial Intelligence},
  volume={39},
number={27},
pages={27858--27868},
year={2025},
doi={10.1609/aaai.v39i27.35002}
}

@inproceedings{lacoste2023geobench,
  author    = {Alexandre Lacoste and Nils Lehmann and Pau Rodriguez and Evan David Sherwin and Hannah Kerner and Bj{\o}rn L{\"u}tjens and Jeremy Andrew Irvin and David Dao and Hamed Alemohammad and Alexandre Drouin and Mehmet Gunturkun and Gabriel Huang and David Vazquez and Dava Newman and Yoshua Bengio and Stefano Ermon and Xiao Xiang Zhu},
  title     = {{GEO-Bench}: Toward Foundation Models for {Earth} Monitoring},
  booktitle = {Advances in Neural Information Processing Systems},
volume = {36},
pages = {51080--51093},
year = {2023},
doi = {10.52202/075280-2223}
}

@inproceedings{
    yeh2021sustainbench,
    title = {{SustainBench}: Benchmarks for Monitoring the {Sustainable Development Goals} with Machine Learning},
    author = {Christopher Yeh and Chenlin Meng and Sherrie Wang and Anne Driscoll and Erik Rozi and Patrick Liu and Jihyeon Lee and Marshall Burke and David Lobell and Stefano Ermon},
    booktitle = {Thirty-fifth Conference on Neural Information Processing Systems, Datasets and Benchmarks Track (Round 2)},
    year = {2021},
    month = {12},
    url = {https://openreview.net/forum?id=5HR3vCylqD}
}

@misc{ESACCI,
  doi = {10.5285/95913FFB6467447CA72C4E9D8CF30501},
  url = {https://catalogue.ceda.ac.uk/uuid/95913ffb6467447ca72c4e9d8cf30501},
  author = {Santoro,  Maurizio and Cartus,  Oliver},
  language = {en},
  title = {{ESA} {Biomass Climate Change Initiative} ({Biomass\_cci}): Global datasets of forest above-ground biomass for the years 2007, 2010, 2015, 2016, 2017, 2018, 2019, 2020, 2021 and 2022, v6.0},
  publisher = {NERC EDS Centre for Environmental Data Analysis},
  year = {2025},
  copyright = {Under the following licence https://artefacts.ceda.ac.uk/licences/specific_licences/esacci_biomass_terms_and_conditions_v2.pdf,  appropriate use of these data may fall under any use. This message is intended as guidance,  always read the full licence. When using these data you must cite them correctly using the citation given on the CEDA Data Catalogue record.}
}

@misc{AGBref,
  doi = {10.5281/ZENODO.15495068},
  url = {https://zenodo.org/doi/10.5281/zenodo.15495068},
  author = {Araza, Arnan and Bielig, Cara and Flores, Ren Avell Anne},
  title = {{AGBref} – Global Reference Dataset for Above-Ground Biomass ({AGB})},
  publisher = {Zenodo},
  year = {2025},
  copyright = {Creative Commons Attribution 4.0 International}
}

@misc{GEDIL4A,
  doi = {10.3334/ORNLDAAC/2056},
  url = {https://www.earthdata.nasa.gov/data/catalog/ornl-cloud-gedi-l4a-agb-density-v2-1-2056-2.1},
  author = {Dubayah,  R.O. and Armston,  J. and Kellner,  J.R. and Duncanson,  L. and Healey,  S.P. and Patterson,  P.L. and Hancock,  S. and Tang,  H. and Bruening,  J.M. and Hofton,  M.A. and Blair,  J.B. and Luthcke,  S.B.},
  language = {en},
  title = {{GEDI} {L4A} Footprint Level Aboveground Biomass Density,  Version 2.1},
  publisher = {ORNL Distributed Active Archive Center},
  year = {2022}
}

@misc{hu2022lowrank,
  doi = {10.48550/ARXIV.2106.09685},
  url = {https://arxiv.org/abs/2106.09685},
  author = {Hu,  Edward J. and Shen,  Yelong and Wallis,  Phillip and Allen-Zhu,  Zeyuan and Li,  Yuanzhi and Wang,  Shean and Wang,  Lu and Chen,  Weizhu},
  title = {{LoRA}: Low-Rank Adaptation of Large Language Models},
  publisher = {arXiv},
  year = {2021},
  copyright = {arXiv.org perpetual,  non-exclusive license}
}

@Article{isprs-annals-II-4-71-2014,
AUTHOR = {Tadono, T. and Ishida, H. and Oda, F. and Naito, S. and Minakawa, K. and Iwamoto, H.},
TITLE = {Precise Global {DEM} Generation by {ALOS PRISM}},
JOURNAL = {ISPRS Annals of the Photogrammetry, Remote Sensing and Spatial Information Sciences},
VOLUME = {II-4},
YEAR = {2014},
PAGES = {71--76},
URL = {https://isprs-annals.copernicus.org/articles/II-4/71/2014/},
DOI = {10.5194/isprsannals-II-4-71-2014}
}

@misc{ESALC,
  doi = {10.5281/ZENODO.3939050},
  url = {https://zenodo.org/record/3939050},
  author = {Buchhorn,  Marcel and Smets,  Bruno and Bertels,  Luc and Roo,  Bert De and Lesiv,  Myroslava and Tsendbazar,  Nandin-Erdene and Herold,  Martin and Fritz,  Steffen},
  language = {en},
  title = {{Copernicus Global Land Service}: Land Cover 100m: collection 3: epoch 2019: Globe},
  publisher = {Zenodo},
  year = {2020},
  copyright = {Creative Commons Attribution 4.0 International}
}

@InProceedings{perez2018film,
  title={{FiLM}: Visual Reasoning with a General Conditioning Layer},
  author={Ethan Perez and Florian Strub and Harm de Vries and Vincent Dumoulin and Aaron C. Courville},
  booktitle={Proceedings of the AAAI Conference on Artificial Intelligence},
year={2018},
doi={10.1609/aaai.v32i1.11671}
}

@InProceedings{Xiao2018,
author = {Xiao, Tete and Liu, Yingcheng and Zhou, Bolei and Jiang, Yuning and Sun, Jian},
title = {Unified Perceptual Parsing for Scene Understanding},
booktitle = {Proceedings of the European Conference on Computer Vision (ECCV)},
pages = {432--448},
month = {September},
year = {2018},
doi = {10.1007/978-3-030-01228-1_26}
}

@article{Potapov2021,
  title = {Mapping global forest canopy height through integration of {GEDI} and {Landsat} data},
  volume = {253},
  ISSN = {0034-4257},
  url = {http://dx.doi.org/10.1016/j.rse.2020.112165},
  DOI = {10.1016/j.rse.2020.112165},
  journal = {Remote Sensing of Environment},
  publisher = {Elsevier BV},
  author = {Potapov,  Peter and Li,  Xinyuan and Hernandez-Serna,  Andres and Tyukavina,  Alexandra and Hansen,  Matthew C. and Kommareddy,  Anil and Pickens,  Amy and Turubanova,  Svetlana and Tang,  Hao and Silva,  Carlos Edibaldo and Armston,  John and Dubayah,  Ralph and Blair,  J. Bryan and Hofton,  Michelle},
  year = {2021},
  month = feb,
  pages = {112165}
}

@misc{Pauls2024,
  doi = {10.48550/ARXIV.2406.01076},
  url = {https://arxiv.org/abs/2406.01076},
  author = {Pauls,  Jan and Zimmer,  Max and Kelly,  Una M. and Schwartz,  Martin and Saatchi,  Sassan and Ciais,  Philippe and Pokutta,  Sebastian and Brandt,  Martin and Gieseke,  Fabian},
  title = {Estimating Canopy Height at Scale},
  publisher = {arXiv},
  year = {2024},
  copyright = {arXiv.org perpetual,  non-exclusive license}
}

@inproceedings{
    pauls2025capturing,
    title={Capturing Temporal Dynamics in Large-Scale Canopy Tree Height Estimation},
    author={Jan Pauls and Max Zimmer and Berkant Turan and Sassan Saatchi and Philippe Ciais and Sebastian Pokutta and Fabian Gieseke},
    booktitle={Forty-second International Conference on Machine Learning},
    year={2025},
    url={https://openreview.net/forum?id=ri1cs3vtXX}
}

@misc{pauls2026echosat,
  doi = {10.48550/ARXIV.2602.21421},
  url = {https://arxiv.org/abs/2602.21421},
  author = {Pauls,  Jan and Schrödter,  Karsten and Ligensa,  Sven and Schwartz,  Martin and Turan,  Berkant and Zimmer,  Max and Saatchi,  Sassan and Pokutta,  Sebastian and Ciais,  Philippe and Gieseke,  Fabian},
  title = {ECHOSAT: Estimating Canopy Height Over Space And Time},
  publisher = {arXiv},
  year = {2026},
  copyright = {Creative Commons Attribution 4.0 International}
}

@article{forgaard2026thor,
      title={{THOR}: A Versatile Foundation Model for {Earth} Observation Climate and Society Applications}, 
      author={Theodor Forgaard and Jarle H. Reksten and Anders U. Waldeland and Valerio Marsocci and Nicolas Longépé and Michael Kampffmeyer and Arnt-Børre Salberg},
      journal={arXiv preprint arXiv:2601.16011},
      year={2026},
      eprint={2601.16011},
      archivePrefix={arXiv},
      primaryClass={eess.IV},
      url={https://arxiv.org/abs/2601.16011}, 
}

@article{SANTORO2026112536,
title = {{Europe}-wide maps of biomass density based on satellite remote sensing data for 2017, 2020, 2021 and 2023},
journal = {Data in Brief},
volume = {65},
pages = {112536},
year = {2026},
issn = {2352-3409},
doi = {10.1016/j.dib.2026.112536},
url = {https://www.sciencedirect.com/science/article/pii/S2352340926000892},
author = {Maurizio Santoro and Oliver Cartus and Arnan Araza and Martin Herold and Jukka Miettinen and Ake Rosenqvist and Kazufumi Kobayashi and Takeo Tadono and Frank Martin Seifert}
}

@article{Jia2023,
  title = {Accuracy evaluation and effect factor analysis of {GEDI} aboveground biomass product for temperate forests in the conterminous {United States}},
  volume = {61},
  ISSN = {1943-7226},
  url = {http://dx.doi.org/10.1080/15481603.2023.2292374},
  DOI = {10.1080/15481603.2023.2292374},
  number = {1},
  journal = {GIScience \& Remote Sensing},
  publisher = {Informa UK Limited},
  author = {Jia,  Duo and Wang,  Cangjiao and Hakkenberg,  Christopher R. and Numata,  Izaya and Elmore,  Andrew J. and Cochrane,  Mark A.},
  year = {2023},
  month = Dec 
}

@article{Pascual2023,
  title = {Assessing the performance of {NASA}’s {GEDI} {L4A} footprint aboveground biomass density models using {National Forest Inventory} and airborne laser scanning data in {Mediterranean} forest ecosystems},
  volume = {538},
  ISSN = {0378-1127},
  url = {http://dx.doi.org/10.1016/j.foreco.2023.120975},
  DOI = {10.1016/j.foreco.2023.120975},
  journal = {Forest Ecology and Management},
  publisher = {Elsevier BV},
  author = {Pascual,  Adrián and Guerra-Hernández,  Juan and Armston,  John and Minor,  David M. and Duncanson,  Laura I. and May,  Paul B. and Kellner,  James R. and Dubayah,  Ralph},
  year = {2023},
  month = Jun,
  pages = {120975}
}

@article{Li2024,
  title = {Evaluation of {GEDI} footprint level biomass models in {Southern African} Savannas using airborne {LiDAR} and field measurements},
  volume = {10},
  ISSN = {2666-0172},
  url = {http://dx.doi.org/10.1016/j.srs.2024.100161},
  DOI = {10.1016/j.srs.2024.100161},
  journal = {Science of Remote Sensing},
  publisher = {Elsevier BV},
  author = {Li,  Xiaoxuan and Wessels,  Konrad and Armston,  John and Duncanson,  Laura and Urbazaev,  Mikhail and Naidoo,  Laven and Mathieu,  Renaud and Main,  Russell},
  year = {2024},
  month = Dec,
  pages = {100161}
}

@article{Naidoo2015,
  title = {Savannah woody structure modelling and mapping using multi-frequency ({X}-,  {C}- and {L}-band) Synthetic Aperture Radar data},
  volume = {105},
  ISSN = {0924-2716},
  url = {http://dx.doi.org/10.1016/j.isprsjprs.2015.04.007},
  DOI = {10.1016/j.isprsjprs.2015.04.007},
  journal = {ISPRS Journal of Photogrammetry and Remote Sensing},
  publisher = {Elsevier BV},
  author = {Naidoo,  Laven and Mathieu,  Renaud and Main,  Russell and Kleynhans,  Waldo and Wessels,  Konrad and Asner,  Gregory and Leblon,  Brigitte},
  year = {2015},
  month = Jul,
  pages = {234–250}
}

@article{Imhoff1995,
  title = {Radar backscatter and biomass saturation: ramifications for global biomass inventory},
  volume = {33},
  ISSN = {1558-0644},
  url = {http://dx.doi.org/10.1109/TGRS.1995.8746034},
  DOI = {10.1109/tgrs.1995.8746034},
  number = {2},
  journal = {IEEE Transactions on Geoscience and Remote Sensing},
  publisher = {Institute of Electrical and Electronics Engineers (IEEE)},
  author = {Imhoff,  Marc L.},
  year = {1995},
  month = Mar,
  pages = {511–518}
}

@article{pan2024,
  title   = {The enduring world forest carbon sink},
  author  = {Pan, Yude and Birdsey, Richard A. and Phillips, Oliver L. and Houghton, Richard A. and Fang, Jingyun and Kauppi, Pekka E. and Keith, Heather and Kurz, Werner A. and Ito, Akihiko and Lewis, Simon L. and Nabuurs, Gert-Jan and Shvidenko, Anatoly and Hashimoto, Shoji and Lerink, Bas and Schepaschenko, Dmitry and Castanho, Andrea and Murdiyarso, Daniel},
  journal = {Nature},
  volume  = {631},
  pages   = {563--569},
  year    = {2024},
  doi     = {10.1038/s41586-024-07602-x}
}

@techreport{FAOUNEP2020,
  title       = {The State of the World's Forests 2020. Forests, biodiversity and people},
  author      = {{FAO} and {UNEP}},
  institution = {FAO and UNEP},
  year        = {2020},
  doi         = {10.4060/ca8642en}
}

@techreport{picard2012,
  title       = {Manual for building tree volume and biomass allometric equations: from field measurement to prediction},
  author      = {Picard, Nicolas and Saint-Andr{\'e}, Laurent and Henry, Matieu},
  institution = {Food and Agriculture Organization of the United Nations (FAO)},
  year        = {2012}
}

@article{chave2014,
  title   = {Improved allometric models to estimate the aboveground biomass of tropical trees},
  author  = {Chave, J{\'e}r{\^o}me and R{\'e}jou-M{\'e}chain, Maxime and B{\'u}rquez, Alberto and Chidumayo, Emmanuel and Colgan, Matthew S. and Delitti, Welington B. C. and Duque, Alvaro and Eid, Tron and Fearnside, Philip M. and Goodman, Rosa C. and Henry, Matieu and Martínez-Yrízar, Angelina and Mugasha, Wilson A. and Muller-Landau, Helene C. and Mencuccini, Maurizio and Nelson, Bruce W. and Ngomanda, Alfred and Nogueira, Euler M. and Ortiz-Malavassi, Edgar and Pélissier, Raphaël and Ploton, Pierre and Ryan, Casey M. and Saldarriaga, Juan G. and Vieilledent, Ghislain},
  journal = {Global Change Biology},
  volume  = {20},
  number  = {10},
  pages   = {3177--3190},
  year    = {2014},
  doi     = {10.1111/gcb.12629}
}

@article{henry2013,
  title   = {{GlobAllomeTree}: international platform for tree allometric equations to support volume, biomass and carbon assessment},
  author  = {Henry, Matieu and Bombelli, Antonio and Trotta, Carlo and Alessandrini, Andrea and Birigazzi, Luca and Sola, Gael and Vieilledent, Ghislain and Santenoise, Philippe and Longuetaud, Fleur and Valentini, Riccardo and Picard, Nicolas and Saint-Andr{\'e}, Laurent},
  journal = {iForest - Biogeosciences and Forestry},
  volume  = {6},
  pages   = {326--330},
  year    = {2013},
  doi     = {10.3832/ifor0901-006}
}

@article{jucker2017,
  title   = {Allometric equations for integrating remote sensing imagery into forest monitoring programmes},
  author  = {Jucker, Tommaso and Caspersen, John and Chave, J{\'e}r{\^o}me and Antin, C{\'e}cile and Barbier, Nicolas and Bongers, Frans and Dalponte, Michele and van Ewijk, Karin Y. and Forrester, David I. and Haeni, Matthias and Higgins, Steven I. and Holdaway, Robert J. and Iida, Yoshiko and Lorimer, Craig and Marshall, Peter L. and Momo, Stéphane and Moncrieff, Glenn R. and Ploton, Pierre and Poorter, Lourens and Rahman, Kassim Abd and Schlund, Michael and Sonké, Bonaventure and Sterck, Frank J. and Trugman, Anna T. and Usoltsev, Vladimir A. and Vanderwel, Mark C. and Waldner, Peter and Wedeux, Beatrice M. M. and Wirth, Christian and Wöll, Hannsjörg and Woods, Murray and Xiang, Wenhua and Zimmermann, Niklaus E. and Coomes, David A.},
  journal = {Global Change Biology},
  volume  = {23},
  number  = {1},
  pages   = {177--190},
  year    = {2017},
  doi     = {10.1111/gcb.13388}
}

@techreport{moges2010,
  title       = {Manual for measurement, monitoring and reporting of carbon stocks in forests and other land uses in Ethiopia},
  author      = {Moges, Yitebitu and Eshetu, Zewdu and Nune, Sisay and Ababa, Addis},
  institution = {UN-REDD Programme},
  year        = {2010}
}

@article{kangas2018,
  title   = {Remote sensing and forest inventories in {Nordic} countries -- roadmap for the future},
  author  = {Kangas, Annika and Astrup, Rasmus and Breidenbach, Johannes and Fridman, Jonas and Gobakken, Terje and Korhonen, Kari T. and Maltamo, Matti and Nilsson, Mats and Nord-Larsen, Thomas and N{\ae}sset, Erik and Olsson, H{\aa}kan},
  journal = {Scandinavian Journal of Forest Research},
  volume  = {33},
  number  = {4},
  pages   = {397--412},
  year    = {2018},
  doi     = {10.1080/02827581.2017.1416666}
}

@article{disney2018,
  title   = {Weighing trees with lasers: advances, challenges and opportunities},
  author  = {Disney, Mathias I. and Boni Vicari, Matheus and Burt, Andrew and Calders, Kim and Lewis, Simon L. and Raumonen, Pasi and Wilkes, Phil},
  journal = {Interface Focus},
  volume  = {8},
  number  = {2},
  pages   = {20170048},
  year    = {2018},
  doi     = {10.1098/rsfs.2017.0048}
}

@article{holvoet2025,
  title   = {Terrestrial and mobile laser scanning for national forest inventories: From theory to implementation},
  author  = {Holvoet, J. and Eichhorn, M. P. and Giannetti, F. and K{\"u}kenbrink, D. and Liang, X. and Mokro{\v{s}}, M. and Novotn{\'y}, J. and Pitk{\"a}nen, T. P. and Puliti, S. and Skudnik, M. and Stere{\'n}czak, K. and Terryn, L. and Vega, C. and Torresan, C.},
  journal = {Remote Sensing of Environment},
  volume  = {329},
  pages   = {114947},
  year    = {2025},
  doi     = {10.1016/j.rse.2025.114947}
}

@article{nilsson2017,
  title   = {A nationwide forest attribute map of {Sweden} predicted using airborne laser scanning data and field data from the National Forest Inventory},
  author  = {Nilsson, Mats and Nordkvist, Karin and Jonz{\'e}n, Jonas and Lindgren, Nils and Axensten, Peder and Wallerman, Jörgen and Egberth, Mikael and Larsson, Sören and Nilsson, Liselott and Eriksson, Jon and Olsson, H{\aa}kan},
  journal = {Remote Sensing of Environment},
  volume  = {194},
  pages   = {447--454},
  year    = {2017},
  doi     = {10.1016/j.rse.2016.10.022}
}

@article{monnet2016,
  title   = {Wide-area mapping of forest with national airborne laser scanning and field inventory datasets},
  author  = {Monnet, J.-M. and Ginzler, C. and Clivaz, J.-C.},
  journal = {The International Archives of the Photogrammetry, Remote Sensing and Spatial Information Sciences},
  volume  = {XLI-B8},
  pages   = {727--731},
  year    = {2016},
  doi     = {10.5194/isprs-archives-XLI-B8-727-2016}
}

@article{drusch2012,
  title   = {{Sentinel-2}: {ESA}'s optical high-resolution mission for {GMES} operational services},
  author  = {Drusch, M. and Del Bello, U. and Carlier, S. and Colin, O. and Fernandez, V. and Gascon, F. and Hoersch, B. and Isola, C. and Laberinti, P. and Martimort, P. and Meygret, A. and Spoto, F. and Sy, O. and Marchese, F. and Bargellini, P.},
  journal = {Remote Sensing of Environment},
  volume  = {120},
  pages   = {25--36},
  year    = {2012},
  doi     = {10.1016/j.rse.2011.11.026}
}

@article{rosenqvist2007,
  title   = {{ALOS PALSAR}: A pathfinder mission for global-scale monitoring of the environment},
  author  = {Rosenqvist, Ake and Shimada, Masanobu and Ito, Nobuhiro and Watanabe, Manabu},
  journal = {IEEE Transactions on Geoscience and Remote Sensing},
  volume  = {45},
  number  = {11},
  pages   = {3307--3316},
  year    = {2007},
  doi     = {10.1109/TGRS.2007.901027}
}

@article{moreira2013,
  title   = {A tutorial on synthetic aperture radar},
  author  = {Moreira, Alberto and Prats-Iraola, Pau and Younis, Marwan and Krieger, Gerhard and Hajnsek, Irena and Papathanassiou, Konstantinos P.},
  journal = {IEEE Geoscience and Remote Sensing Magazine},
  volume  = {1},
  number  = {1},
  pages   = {6--43},
  year    = {2013},
  doi     = {10.1109/MGRS.2013.2248301}
}

@article{dubayah2020,
  title   = {The {Global Ecosystem Dynamics Investigation}: High-resolution laser ranging of the {Earth}'s forests and topography},
  author  = {Dubayah, Ralph and Blair, J. Bryan and Goetz, Scott and Fatoyinbo, Lola and Hansen, Matthew and Healey, Sean and Hofton, Michelle and Hurtt, George and Kellner, James and Luthcke, Scott and Armston, John and Tang, Hao and Duncanson, Laura and Hancock, Steven and Jantz, Patrick and Marselis, Suzanne and Patterson, Paul L. and Qi, Wenlu and Silva, Carlos},
  journal = {Science of Remote Sensing},
  volume  = {1},
  pages   = {100002},
  year    = {2020},
  doi     = {10.1016/j.srs.2020.100002}
}

@InProceedings{Houdre_2026_CVPR,
    author    = {Houdr\'e, Nicolas and Marcos, Diego and de Turckheim, Hugo Riffaud and Ienco, Dino and Wendling, Laurent and Kurtz, Camille and Lobry, Sylvain},
    title     = {{RAMEN}: Resolution-Adjustable Multimodal Encoder for {Earth} Observation},
    booktitle = {Proceedings of the IEEE/CVF Conference on Computer Vision and Pattern Recognition (CVPR)},
    month     = {June},
    year      = {2026},
    pages     = {27838-27848}
}

@inproceedings{Sumbul2025,
  title = {{SMARTIES}: Spectrum-Aware Multi-Sensor Auto-Encoder for Remote Sensing Images},
  url = {http://dx.doi.org/10.1109/ICCV51701.2025.00528},
  DOI = {10.1109/iccv51701.2025.00528},
  booktitle = {Proceedings of the IEEE/CVF International Conference on Computer Vision (ICCV)},
  publisher = {IEEE},
  author = {Sumbul,  Gencer and Xu,  Chang and Dalsasso,  Emanuele and Tuia,  Devis},
  year = {2025},
  month = Oct,
  pages = {5569--5578}
}

@article{wang2022ssl4eo,
  title={{SSL4EO-S12}: A Large-Scale Multimodal, Multitemporal Dataset for Self-Supervised Learning in {Earth} Observation},
  author={Wang, Yi and Braham, Nassim Ait Ali and Xiong, Zhitong and Liu, Chenying and Albrecht, Conrad M. and Zhu, Xiao Xiang},
  journal = {IEEE Geoscience and Remote Sensing Magazine},
volume = {11},
number = {3},
pages = {98--106},
year = {2023},
doi = {10.1109/MGRS.2023.3281651}
}

@article{TOLAN2024113888,
title = {Very high resolution canopy height maps from {RGB} imagery using self-supervised vision transformer and convolutional decoder trained on aerial lidar},
journal = {Remote Sensing of Environment},
volume = {300},
pages = {113888},
year = {2024},
issn = {0034-4257},
doi = {10.1016/j.rse.2023.113888},
url = {https://www.sciencedirect.com/science/article/pii/S003442572300439X},
author = {Jamie Tolan and Hung-I Yang and Benjamin Nosarzewski and Guillaume Couairon and Huy V. Vo and John Brandt and Justine Spore and Sayantan Majumdar and Daniel Haziza and Janaki Vamaraju and Theo Moutakanni and Piotr Bojanowski and Tracy Johns and Brian White and Tobias Tiecke and Camille Couprie}}

@Article{schwartz2023forms,
AUTHOR = {Schwartz, M. and Ciais, P. and De Truchis, A. and Chave, J. and Ottl\'e, C. and Vega, C. and Wigneron, J.-P. and Nicolas, M. and Jouaber, S. and Liu, S. and Brandt, M. and Fayad, I.},
TITLE = {{FORMS}: Forest Multiple Source height, wood volume, and biomass maps in
{France} at 10 to 30\,m resolution based on {Sentinel-1}, {Sentinel-2}, and {Global Ecosystem Dynamics Investigation} ({GEDI}) data with a deep learning approach},
JOURNAL = {Earth System Science Data},
VOLUME = {15},
YEAR = {2023},
NUMBER = {11},
PAGES = {4927--4945},
URL = {https://essd.copernicus.org/articles/15/4927/2023/},
DOI = {10.5194/essd-15-4927-2023}
}

@article{zhu2017deep,
  title={Deep learning in remote sensing: A comprehensive review and list of resources},
  author={Zhu, Xiao Xiang and Tuia, Devis and Mou, Lichao and Xia, Gui-Song and Zhang, Liangpei and Xu, Feng and Fraundorfer, Friedrich},
  journal={IEEE Geoscience and Remote Sensing Magazine},
  volume={5},
  number={4},
  pages={8--36},
  year={2017},
  publisher={IEEE}
}

@article{avitabile2016,
  title = {An integrated pan‐tropical biomass map using multiple reference datasets},
  volume = {22},
  ISSN = {1365-2486},
  url = {http://dx.doi.org/10.1111/gcb.13139},
  DOI = {10.1111/gcb.13139},
  number = {4},
  journal = {Global Change Biology},
  publisher = {Wiley},
  author = {Avitabile,  Valerio and Herold,  Martin and Heuvelink,  Gerard B. M. and Lewis,  Simon L. and Phillips,  Oliver L. and Asner,  Gregory P. and Armston,  John and Ashton,  Peter S. and Banin,  Lindsay and Bayol,  Nicolas and Berry,  Nicholas J. and Boeckx,  Pascal and de Jong,  Bernardus H. J. and DeVries,  Ben and Girardin,  Cecile A. J. and Kearsley,  Elizabeth and Lindsell,  Jeremy A. and Lopez‐Gonzalez,  Gabriela and Lucas,  Richard and Malhi,  Yadvinder and Morel,  Alexandra and Mitchard,  Edward T. A. and Nagy,  Laszlo and Qie,  Lan and Quinones,  Marcela J. and Ryan,  Casey M. and Ferry,  Slik J. W. and Sunderland,  Terry and Laurin,  Gaia Vaglio and Gatti,  Roberto Cazzolla and Valentini,  Riccardo and Verbeeck,  Hans and Wijaya,  Arief and Willcock,  Simon},
  year = {2016},
  month = Jan,
  pages = {1406–1420}
}

@article{RodrguezVeiga2019,
  title = {Forest biomass retrieval approaches from earth observation in different biomes},
  volume = {77},
  ISSN = {1569-8432},
  url = {http://dx.doi.org/10.1016/j.jag.2018.12.008},
  DOI = {10.1016/j.jag.2018.12.008},
  journal = {International Journal of Applied Earth Observation and Geoinformation},
  publisher = {Elsevier BV},
  author = {Rodríguez-Veiga,  Pedro and Quegan,  Shaun and Carreiras,  Joao and Persson,  Henrik J. and Fransson,  Johan E.S. and Hoscilo,  Agata and Ziółkowski,  Dariusz and Stereńczak,  Krzysztof and Lohberger,  Sandra and St\"{a}ngel,  Matthias and Berninger,  Anna and Siegert,  Florian and Avitabile,  Valerio and Herold,  Martin and Mermoz,  Stéphane and Bouvet,  Alexandre and Le Toan,  Thuy and Carvalhais,  Nuno and Santoro,  Maurizio and Cartus,  Oliver and Rauste,  Yrj\"{o} and Mathieu,  Renaud and Asner,  Gregory P. and Thiel,  Christian and Pathe,  Carsten and Schmullius,  Chris and Seifert,  Frank Martin and Tansey,  Kevin and Balzter,  Heiko},
  year = {2019},
  month = May,
  pages = {53–68}
}

@article{Duncanson2022,
  title = {Aboveground biomass density models for {NASA}’s {Global Ecosystem Dynamics Investigation} ({GEDI}) lidar mission},
  volume = {270},
  ISSN = {0034-4257},
  url = {http://dx.doi.org/10.1016/j.rse.2021.112845},
  DOI = {10.1016/j.rse.2021.112845},
  journal = {Remote Sensing of Environment},
  publisher = {Elsevier BV},
  author = {Duncanson,  Laura and Kellner,  James R. and Armston,  John and Dubayah,  Ralph and Minor,  David M. and Hancock,  Steven and Healey,  Sean P. and Patterson,  Paul L. and Saarela,  Svetlana and Marselis,  Suzanne and Silva,  Carlos E. and Bruening,  Jamis and Goetz,  Scott J. and Tang,  Hao and Hofton,  Michelle and Blair,  Bryan and Luthcke,  Scott and Fatoyinbo,  Lola and Abernethy,  Katharine and Alonso,  Alfonso and Andersen,  Hans-Erik and Aplin,  Paul and Baker,  Timothy R. and Barbier,  Nicolas and Bastin,  Jean Francois and Biber,  Peter and Boeckx,  Pascal and Bogaert,  Jan and Boschetti,  Luigi and Boucher,  Peter Brehm and Boyd,  Doreen S. and Burslem,  David F.R.P. and Calvo-Rodriguez,  Sofia and Chave,  Jér\^ome and Chazdon,  Robin L. and Clark,  David B. and Clark,  Deborah A. and Cohen,  Warren B. and Coomes,  David A. and Corona,  Piermaria and Cushman,  K.C. and Cutler,  Mark E.J. and Dalling,  James W. and Dalponte,  Michele and Dash,  Jonathan and de-Miguel,  Sergio and Deng,  Songqiu and Ellis,  Peter Woods and Erasmus,  Barend and Fekety,  Patrick A. and Fernandez-Landa,  Alfredo and Ferraz,  Antonio and Fischer,  Rico and Fisher,  Adrian G. and García-Abril,  Antonio and Gobakken,  Terje and Hacker,  Jorg M. and Heurich,  Marco and Hill,  Ross A. and Hopkinson,  Chris and Huang,  Huabing and Hubbell,  Stephen P. and Hudak,  Andrew T. and Huth,  Andreas and Imbach,  Benedikt and Jeffery,  Kathryn J. and Katoh,  Masato and Kearsley,  Elizabeth and Kenfack,  David and Kljun,  Natascha and Knapp,  Nikolai and Král,  Kamil and Krůček,  Martin and Labrière,  Nicolas and Lewis,  Simon L. and Longo,  Marcos and Lucas,  Richard M. and Main,  Russell and Manzanera,  Jose A. and Martínez,  Rodolfo Vásquez and Mathieu,  Renaud and Memiaghe,  Herve and Meyer,  Victoria and Mendoza,  Abel Monteagudo and Monerris,  Alessandra and Montesano,  Paul and Morsdorf,  Felix and Næsset,  Erik and Naidoo,  Laven and Nilus,  Reuben and O’Brien,  Michael and Orwig,  David A. and Papathanassiou,  Konstantinos and Parker,  Geoffrey and Philipson,  Christopher and Phillips,  Oliver L. and Pisek,  Jan and Poulsen,  John R. and Pretzsch,  Hans and R\"{u}diger,  Christoph and Saatchi,  Sassan and Sanchez-Azofeifa,  Arturo and Sanchez-Lopez,  Nuria and Scholes,  Robert and Silva,  Carlos A. and Simard,  Marc and Skidmore,  Andrew and Stereńczak,  Krzysztof and Tanase,  Mihai and Torresan,  Chiara and Valbuena,  Ruben and Verbeeck,  Hans and Vrska,  Tomas and Wessels,  Konrad and White,  Joanne C. and White,  Lee J.T. and Zahabu,  Eliakimu and Zgraggen,  Carlo},
  year = {2022},
  month = Mar,
  pages = {112845}
}

@article{perron2026universat,
  title   = {{UniverSat}: Resolution- and Modality-Agnostic Transformers for {Earth} Observation},
  author  = {Perron, Yohann and Astruc, Guillaume and Gonthier, Nicolas
             and Mallet, Clement and Landrieu, Loic},
  journal = {arXiv preprint arXiv:2606.23503},
  year    = {2026}
}

@article{Bourriz2026,
  title = {How Effective Are Foundation Models for Crop Type Mapping Using Hyperspectral Imaging? A Comparative Study of Machine Learning,  Deep Learning,  and Geospatial Foundation Models},
  volume = {XLVIII-4/W17-2025},
  ISSN = {2194-9034},
  url = {http://dx.doi.org/10.5194/isprs-archives-XLVIII-4-W17-2025-69-2026},
  DOI = {10.5194/isprs-archives-xlviii-4-w17-2025-69-2026},
  journal = {The International Archives of the Photogrammetry,  Remote Sensing and Spatial Information Sciences},
  publisher = {Copernicus GmbH},
  author = {Bourriz,  Mohamed and Laamrani,  Ahmed and Ait Abdelali,  Hamd and Bourzeix,  Fran\c{c}ois and El-Battay,  Ali and Amazirh,  Abdelhakim and Chehbouni,  Abdelghani},
  year = {2026},
  month = Jan,
  pages = {69–74}
}

@misc{Hossain2026,
  doi = {10.48550/ARXIV.2610.05790},
  url = {https://arxiv.org/abs/2610.05790},
  author = {Hossain,  Md Aminur and Vaghasiya,  Omkumar and Dwivedi,  Rajeev Ranjan and Kurmi,  Vinod and Banerjee,  Biplab},
  title = {{FairRSFM}: A Biome-Aware Benchmark and Debiasing Framework for Remote Sensing Foundation Models},
  publisher = {arXiv},
  year = {2026},
  copyright = {arXiv.org perpetual,  non-exclusive license}
}

@misc{Kaushik2026,
  doi = {10.48550/ARXIV.2603.01576},
  url = {https://arxiv.org/abs/2603.01576},
  author = {Kaushik,  Saurabh and Maurya,  Lalit and Tellman,  Beth and Roy,  Swalpa Kumar and Marsocci,  Valerio and Camps-Valls,  Gustau and Chanussot,  Jocelyn},
  title = {{Cryo-Bench}: Benchmarking Foundation Models for Cryosphere Mapping},
  publisher = {arXiv},
  year = {2026},
  copyright = {Creative Commons Attribution 4.0 International}
}

@article{RojasLucero2026,
  title = {Spectral indices outperform {AlphaEarth} foundation embeddings for aboveground biomass estimation in a regenerating tropical {Andean} forest},
  volume = {42},
  ISSN = {2352-9385},
  url = {http://dx.doi.org/10.1016/j.rsase.2026.102113},
  DOI = {10.1016/j.rsase.2026.102113},
  journal = {Remote Sensing Applications: Society and Environment},
  publisher = {Elsevier BV},
  author = {Rojas-Lucero,  J.C. and Solano-Correa,  Y.T. and Kolarik,  N.E. and Brandt,  J.S. and Caughlin,  T.},
  year = {2026},
  month = Apr,
  pages = {102113}
}

@article{Jin2026,
  title = {Assessing the Utility of Satellite Embedding Features for Biomass Prediction in Subtropical Forests with Machine Learning},
  volume = {18},
  ISSN = {2072-4292},
  url = {http://dx.doi.org/10.3390/rs18030436},
  DOI = {10.3390/rs18030436},
  number = {3},
  journal = {Remote Sensing},
  publisher = {MDPI AG},
  author = {Jin,  Chao and Jiang,  Xiaodong and Wen,  Lina and Wu,  Chuping and Xu,  Xia and Jiao,  Jiejie},
  year = {2026},
  month = Jan,
  pages = {436}
}

@article{Pascual2026,
  title = {Integration of {Google}’s {Alpha Earth Foundations} into biomass estimation combined with {GEDI} spaceborne lidar and field inventory data},
  volume = {606},
  ISSN = {0378-1127},
  url = {http://dx.doi.org/10.1016/j.foreco.2026.123550},
  DOI = {10.1016/j.foreco.2026.123550},
  journal = {Forest Ecology and Management},
  publisher = {Elsevier BV},
  author = {Pascual,  Adrian and Guerra-Hernández,  Juan},
  year = {2026},
  month = Apr,
  pages = {123550}
}

@misc{muszynski2024,
  doi = {10.48550/ARXIV.2406.19888},
  url = {https://arxiv.org/abs/2406.19888},
  author = {Muszynski,  Michal and Klein,  Levente and da Silva,  Ademir Ferreira and Atluri,  Anjani Prasad and Gomes,  Carlos and Szwarcman,  Daniela and Singh,  Gurkanwar and Gu,  Kewen and Zortea,  Maciel and Simumba,  Naomi and Fraccaro,  Paolo and Singh,  Shraddha and Meliksetian,  Steve and Watson,  Campbell and Kimura,  Daiki and Srinivasan,  Harini},
  title = {Fine-tuning of Geospatial Foundation Models for Aboveground Biomass Estimation},
  publisher = {arXiv},
  year = {2024},
  copyright = {arXiv.org perpetual,  non-exclusive license}
}

@article{hamoudzadeh2026,
  title   = {Inferring Height from {Earth} Embeddings: First Insights using {Google} {AlphaEarth}},
  author  = {Hamoudzadeh, Alireza and Belloni, Valeria and Ravanelli, Roberta},
  journal = {arXiv preprint arXiv:2602.17250},
  year    = {2026}
}

@article{ma2026harvesting,
  title   = {Harvesting {AlphaEarth}: Benchmarking the geospatial foundation model for agricultural downstream tasks},
  author  = {Ma, Yuchi and Shen, Yawen and Swatantran, Anu and Lobell, David B.},
  journal = {International Journal of Applied Earth Observation and Geoinformation},
  volume  = {149},
  pages   = {105258},
  year    = {2026},
  doi     = {10.1016/j.jag.2026.105258}
}

@article{yang2026irrigated,
  title   = {Evaluating the Performance of {AlphaEarth} Foundation Embeddings for Irrigated Cropland Mapping Across Regions and Years},
  author  = {Yang, Lulu and Gao, Yuan and Zhao, Xiangyang and Liang, Nannan and Ma, Ru and Xi, Shixiang and Zhang, Xiao and Wang, Rui},
  journal = {Remote Sensing},
  volume  = {18},
  number  = {7},
  pages   = {1065},
  year    = {2026},
  doi     = {10.3390/rs18071065}
}


\clearpage
\renewcommand{\listfigurename}{List of Figure Captions}
\phantomsection
\addcontentsline{toc}{section}{\listfigurename}
\listoffigures

\end{document}